\documentclass[10pt,twocolumn,letterpaper]{article}

\usepackage[pagenumbers]{cvpr}

\definecolor{BenchGray}{HTML}{F0F0F2}
\definecolor{BenchMeta}{HTML}{555555}
\definecolor{OursBlue}{HTML}{E8EEF9}
\definecolor{GenBg}{HTML}{D8E5FB}
\definecolor{GenFg}{HTML}{184A91}
\definecolor{RealBg}{HTML}{DDEED0}
\definecolor{RealFg}{HTML}{285E22}
\definecolor{CaptureBg}{HTML}{E8DCF3}
\definecolor{CaptureFg}{HTML}{66338D}
\newcommand{\PhysicsPriorTag}[3]{%
  \tikz[baseline=(tag.base)]{\node[fill=#1,text=#2,rounded corners=5pt,
    inner xsep=7pt,inner ysep=1pt,outer sep=0pt,font=\small\bfseries] (tag) {#3};}%
}
\newcommand{\PhysicsBenchRow}[6]{%
  \rowcolor{BenchGray}
  \multicolumn{7}{p{\dimexpr\linewidth-2\tabcolsep\relax}}{%
    \rule[-3pt]{0pt}{14pt}%
    \makebox[\linewidth][l]{\makebox[64pt][l]{\textbf{#1}}%
      \PhysicsPriorTag{#3}{#4}{#2}\hfill
      {\footnotesize\itshape\color{BenchMeta}#5 assets / #6 trajectories}}}\\[1pt]
}

\definecolor{cvprblue}{rgb}{0.21,0.49,0.74}
\usepackage[pagebackref,breaklinks,colorlinks,allcolors=cvprblue]{hyperref}
\def\confName{CVPR}
\def\confYear{2027}

\usepackage{amsmath,amsfonts,bm}

\def\eqref#1{equation~\ref{#1}}

\def\1{\bm{1}}

\DeclareMathAlphabet{\mathsfit}{\encodingdefault}{\sfdefault}{m}{sl}
\SetMathAlphabet{\mathsfit}{bold}{\encodingdefault}{\sfdefault}{bx}{n}

\usepackage{booktabs}
\usepackage{multirow}
\usepackage{colortbl}
\usepackage{tikz}
\usepackage{float}
\usepackage{placeins}
\usepackage{makecell}
\usepackage{tabularx}
\usepackage{adjustbox}
\usepackage[most]{tcolorbox}
\usepackage{enumitem}
\usepackage{fontawesome5}
\definecolor{promptbg}{RGB}{245, 247, 250}
\definecolor{promptframe}{RGB}{60, 80, 100}
\definecolor{alertred}{RGB}{180, 30, 30}
\newcommand{\SectionTitle}[1]{\par\noindent{\rmfamily\fontsize{10.5}{12}\selectfont\bfseries #1}\par\vspace{2pt}}
\newcommand{\SubTitle}[1]{\noindent{\rmfamily\bfseries\color{promptframe}#1}}

\usetikzlibrary{arrows.meta,calc,positioning}
\usepackage[capitalize]{cleveref}

\crefname{section}{Sec.}{Secs.}
\Crefname{section}{Section}{Sections}
\Crefname{table}{Table}{Tables}
\crefname{table}{Tab.}{Tabs.}

\newcommand{\om}{Art2Sim}
\newcommand{\projecturl}{https://alvinyh.github.io/Art2Sim}

\title{Art2Sim: Learning Physically Executable Full-Body Humanoid Interaction with Articulated Objects from Video Priors}

\author{%
Hang Ye$^{1,*}$\quad Yuan Xu$^{1,*}$\quad Baiting Zhan$^{2,*}$\quad Jinlu Zhang$^{1}$\\[2pt]
Tianshu Wu$^{1}$\quad Qin Xu$^{3}$\quad Hao Dong$^{1}$\quad Yizhou Wang\textsuperscript{1,\scalebox{0.6}{\faEnvelope[regular]}}\\[7pt]
$^{1}$CFCS, School of Computer Science, Peking University\\[2pt]
$^{2}$Yuanpei College, Peking University\quad $^{3}$School of EECS, Peking University%
}

\begin{document}

\makeatletter
\patchcmd{\@maketitle}{\vspace*{24pt}}{\vspace*{12pt}}{}{}
\g@addto@macro\@maketitle{%
  \begin{center}
   \vspace{-0.8em}
    \includegraphics[width=0.95\textwidth]{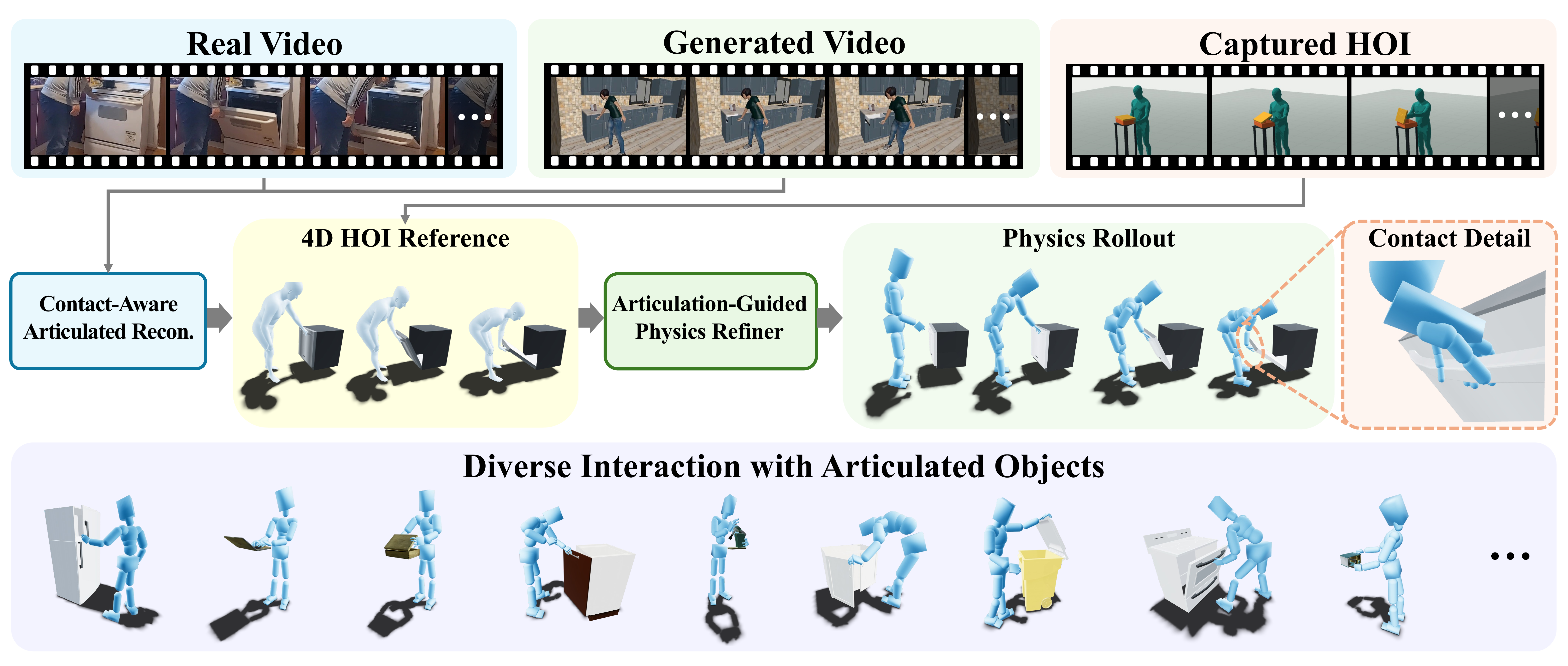}
    \vspace{-0.8em}
    \captionof{figure}{\textbf{\om{} converts diverse interaction priors into physically executable whole-body interactions with articulated objects.} Across real videos, asset-conditioned generated videos, and captured HOI references, the humanoid maintains contact with the moving link and drives the passive joint in simulation. The inset zooms in on the hand-link contact.}
    \vspace{-0.2em}
    \label{fig:teaser}
  \end{center}
  \vspace{1.0em}
}
\makeatother

\maketitle
\begingroup
\renewcommand\thefootnote{}%
\renewcommand\theHfootnote{equal}%
\footnote{$^*$Equal contribution. \quad \textsuperscript{\scalebox{0.65}{\faEnvelope[regular]}}Corresponding author.}%
\addtocounter{footnote}{-1}%
\endgroup

\begin{abstract}
Videos offer a scalable source of whole-body human-object interactions, yet visually coherent trajectories recovered from them do not guarantee successful physical execution.
This gap is especially acute for articulated objects, where a humanoid must sustain contact with the correct moving link and coordinate its whole body to drive the passive joint.
We present \textbf{\om{}}, a complete video-to-physics pipeline that turns monocular interaction videos paired with articulated assets into physically executable humanoid skills.
Contact-Aware Articulated Reconstruction tracks the articulation at the trajectory level and aligns the arm and hand with the moving link using vision-language contact cues and a grasp prior on the link surface.
It recovers more accurate hand-link contact and smoother motion than prior articulated HOI reconstruction methods, while optimizing nearly $10\times$ faster on generated videos.
Articulation-Guided Physics Refinement then learns a residual policy on top of a frozen pretrained whole-body tracker, allowing each per-asset policy to focus on contact and actuation across diverse trajectories of that object.
Fine-grained contact and phase-grounded progress rewards keep the fingers on the moving link and pace the passive joint along the reference.
On real videos, \om{} improves execution success from 38.6\% for the strongest physics-based HOI baseline to 68.1\%, while also leading on generated videos and motion-captured references.
Moreover, under the same physics refiner, our reconstructions consistently outperform those of the strongest reconstruction baseline.
Preliminary experiments on policy distillation and humanoid-robot retargeting suggest potential for downstream embodied applications.~Project~Page:~\href{\projecturl}{\projecturl}
\end{abstract}

\section{Introduction}
\label{sec:introduction}

Articulated objects such as doors, drawers, and cabinets are ubiquitous in human environments, and manipulating them is a core capability for humanoid robots~\citep{jiang2023chairs,weng2025hdmi}.
Learning diverse articulated interactions requires demonstrations across objects and behaviors, but motion capture and teleoperation are expensive to collect at scale.
Monocular videos provide widely available human-object interaction priors, yet existing methods mainly recover visually plausible kinematic motion without ensuring physical feasibility~\citep{xu2021d3dhoi,wang2026arthoi,huang2026arthoi}.
In parallel, physics-based HOI methods learn full-body interaction policies from motion references~\citep{wang2023physhoi,wang2025skillmimic,xu2025intermimic,xu2026interprior}, while recent efforts on full-body articulated manipulation still rely on motion-captured object poses or annotated contact trajectories~\citep{weng2025hdmi}.
Recent work further bridges monocular videos and physics-based control by reproducing reconstructed interactions in simulation~\citep{xie2026grail,huang2026repho,wu2026sugar,kim2026devi}, but remains largely confined to rigid objects.
Articulated objects introduce a distinct challenge because joint structures restrict object motion, requiring the humanoid to establish contact with the correct moving link and generate forces that drive the passive joint along its constrained degrees of freedom.
Converting monocular articulated HOI videos directly into physically executable full-body humanoid policies remains an open problem.

Physically executing articulated interactions from monocular video requires three capabilities beyond motion tracking~\citep{zhang2024artigrasp,deng2026vlmrmd,zhang2026dragmesh2,zhu2026chord}: (1) \emph{contact-aware reference recovery} to overcome occlusion and depth ambiguity, coherently tracking part articulation while aligning the hand with the moving contact surface from limited observations; (2) \emph{contact-mediated joint actuation} to realize this hand-link relation under contact dynamics and generate forces that advance the passive joint; and (3) \emph{sustained whole-body coordination} to continually adjust posture and balance as the link moves (\eg, bending the body while following a rotating low oven door in~\cref{fig:teaser}).

To address this challenge, we introduce \textbf{\om{}}, a video-to-physics framework bridging contact-aware reconstruction and physics execution via a shared contact reference (\cref{fig:teaser}).
Given a monocular video and a paired articulated asset, \textbf{Contact-Aware Articulated Reconstruction} tracks full-sequence object articulation at the trajectory level and refines the arm and hand poses to align with the moving contact surface, recovering smoother motion and more accurate contact nearly $10\times$ faster than frame-by-frame baselines.
Next, \textbf{Articulation-Guided Physics Refinement} learns a closed-loop residual policy atop a frozen dexterous whole-body tracker~\citep{luo2024omnigrasp}.
Guided by fine-grained contact and phase-grounded progress rewards, the policy learns to sustain actionable finger contact and actuate the passive joint across diverse trajectories of each asset.

We evaluate \om{} on real videos from D3D-HOI~\citep{xu2021d3dhoi}, synthesized RoboCasa videos~\citep{nasiriany2024robocasa}, and captured ARCTIC references~\citep{fan2023arctic}.
\om{} recovers superior hand-link contact at a fraction of the optimization cost, which directly translates into higher physics execution success under the same refiner.
Under matched RL training budgets, our per-asset policies raise execution success on real videos from $38.6\%$ for the strongest baseline to $68.1\%$, while achieving the highest success on both generated and captured benchmarks.
The resulting physically executable interactions further support goal-conditioned policy distillation and cross-embodiment humanoid retargeting.
In summary, our contributions are:
\begin{itemize}
\item \textbf{\om{}}, a scalable video-to-physics framework that converts monocular articulated HOI videos into physically executable full-body humanoid skills.
\item A two-stage pipeline combining efficient trajectory-level reconstruction and contact-aware HOI alignment with a residual RL policy guided by fine-grained contact and phase-grounded progress rewards for physics refinement.
\item Comprehensive validation across real, generated, and captured interactions plus downstream transfer, achieving nearly $10\times$ faster reconstruction on generated videos and substantial gains in physical execution success.
\end{itemize}
\section{Related Work}
\label{sec:related-work}

\noindent\textbf{Articulated object and interaction reconstruction.}
Object-centric methods recover part geometry and joint parameters from images
and videos~\citep{qian2022understanding,liu2025videoartgs}, while
D3D-HOI~\citep{xu2021d3dhoi} incorporates human motion cues for articulated
object reconstruction. Hand-centric methods study bimanual articulated
interaction reconstruction~\citep{fan2023arctic,bansal2026hopformer,wang2026arthoi},
and full-body approaches typically rely on multi-view or motion-capture
inputs~\citep{jiang2023chairs,kim2025parahome}. Monocular world-frame
reconstruction has mainly focused on rigid objects~\citep{xue2026rhino}.
However, these approaches focus on kinematic plausibility, leaving physical executability unverified.

\vspace{0.5ex}

\noindent\textbf{Kinematic HOI generation.}
Kinematic HOI generation methods synthesize human-object interactions through
motion generation or optimization. Existing approaches study whole-body
grasping and contact control~\citep{taheri2022goal,zhang2022couch}, diffusion
models with physics-informed corrections or contact
guidance~\citep{xu2023interdiff,li2024chois}, and task-driven interactions with
rigid props~\citep{hassan2023synthesizing}.
For articulated objects, CoDA~\citep{pi2025coda} and GIRAF~\citep{zhang2026giraf} synthesize whole-body manipulation by explicitly generating object joint trajectories rather than driving them through physical contact dynamics.

\vspace{0.5ex}

\noindent\textbf{Physics-based HOI.}
Physics-based HOI methods learn humanoid policies that execute interaction
references under physical dynamics. Humanoid tracking has evolved from motion
imitation~\citep{peng2018deepmimic,peng2021amp} to robust tracking of noisy or
partial references~\citep{yuan2021simpoe,luo2023perpetual,luo2024universal,tessler2024maskedmimic}.
For object interaction, PhysHOI~\citep{wang2023physhoi} introduces
contact-aware imitation rewards, which are extended to broader skills and
object diversity~\citep{wang2025skillmimic,luo2024omnigrasp,xu2025intermimic}.
Recent works further bridge reference generation and humanoid execution through
video-driven reconstruction or residual
control~\citep{xie2026grail,huang2026repho,wu2026sugar,zhao2025resmimic}.
InterPrior~\citep{xu2026interprior} additionally studies goal-driven
whole-body object interaction using a generative prior.
Still, most controllers target rigid objects, leaving contact-driven actuation of reconstructed articulated objects largely open.

\vspace{0.5ex}

\noindent\textbf{Contact-driven articulated manipulation.}
Contact-driven articulation has been studied with dexterous hands, robot
grippers, and demonstration
retargeting~\citep{zhang2024artigrasp,zhang2026dragmesh2,mandi2026dexmachina,kerr2024rsrd}.
While showing that contact can drive articulation, they mostly target disembodied hands or specialized embodiments.
Recent works further investigate contact modeling:
CHORD~\citep{zhu2026chord} reproduces demonstrated object-level contact
wrenches, while WristMimic~\citep{yu2026wristmimic} uses kinematic wrist
guidance without force- or torque-level supervision.
In contrast, turning monocular full-body videos into contact-driven articulated humanoid skills remains unaddressed.

\section{Method}
\label{sec:method}

\begin{figure*}[t]
  \centering
  \includegraphics[width=0.92\textwidth]{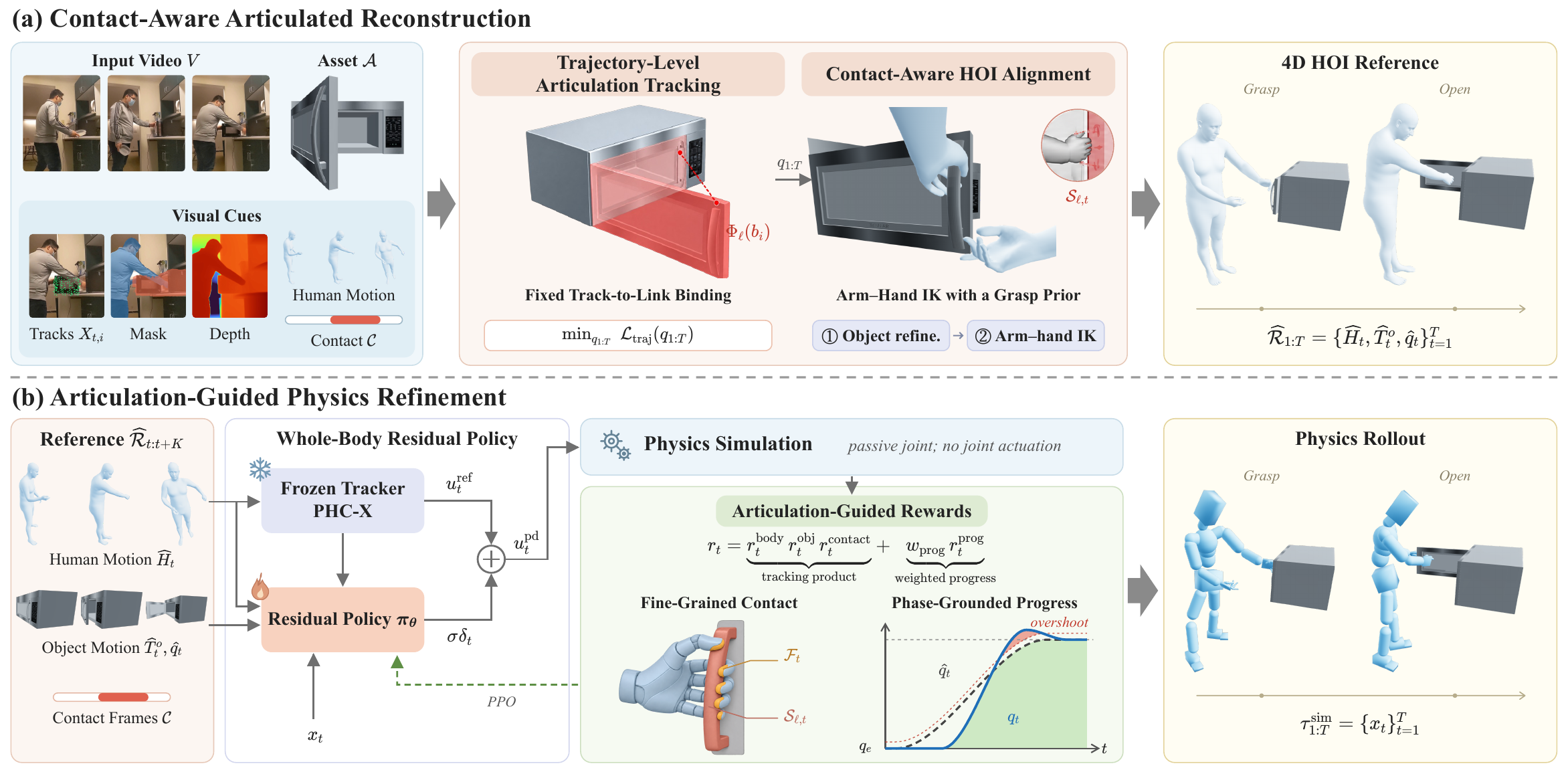}
  \vspace{-1ex}
  \caption{\textbf{Overview of \om{}.} (a) \textbf{Contact-Aware Articulated Reconstruction} first extracts visual cues from a monocular video. It recovers the articulation at the trajectory level and aligns the arms and hands with the contact surface of the moving part, producing a 4D HOI reference. (b) \textbf{Articulation-Guided Physics Refinement} adds a learned residual to a frozen whole-body tracker and trains the residual policy with articulation-aware rewards, producing a physics rollout that drives the object's passive joint through contact.}
  \vspace{-0.4em}
  \label{fig:overview}
\end{figure*}

\om{} converts a monocular human-object interaction video and a paired articulated asset into a physically executable full-body humanoid skill (\cref{fig:overview}).
To bridge the gap between noisy visual observations and contact-rich articulated dynamics, our framework operates in two coupled stages.
First, \textbf{Contact-Aware Articulated Reconstruction} recovers a 4D kinematic reference by efficient trajectory-level articulation tracking and contact-aware HOI alignment (\cref{sec:contact-aware-reconstruction}).
Second, \textbf{Articulation-Guided Physics Refinement} adapts a pretrained whole-body motion tracker through residual reinforcement learning guided by fine-grained contact and articulation-progress rewards, enabling the humanoid to sustain contact and actuate the passive joint in simulation (\cref{sec:articulation-guided-refiner}).

\subsection{Problem Formulation}
\label{sec:problem-formulation}

Given a $T$-frame monocular video $V=\{I_t\}_{t=1}^{T}$ and a matched articulated object asset $\mathcal{A}$ specifying canonical geometry and joint kinematics, our goal is to reconstruct the observed interaction and reproduce it with a full-body humanoid in physics simulation.
In practice, $\mathcal{A}$ is either the 3D asset used to condition a generated video or a matched articulated CAD model of the object in a real video.
The reconstruction stage maps $(V,\mathcal{A})$ to a metric object scale $\widehat{s}\in\mathbb{R}_{+}$ and a 4D kinematic reference:
\begin{equation}
\widehat{\mathcal{R}}_{1:T}
= \operatorname{Recon}(V, \mathcal{A})
= \big\{\widehat{H}_t, \widehat{T}^{o}_t, \widehat{q}_t\big\}_{t=1}^{T},
\label{eq:reconstruction-reference}
\end{equation}
where $\widehat{H}_t$ denotes the full-body SMPL-X human motion, $\widehat{T}^{o}_t \in \mathrm{SE}(3)$ is the object root pose, and $\widehat{q}_t \in \mathbb{R}^{J}$ represents the articulated joint states.
Conditioned on $\widehat{\mathcal{R}}_{1:T}$, the physics refiner learns a closed-loop humanoid policy $\pi_{\theta}(a_t \mid x_t,\widehat{\mathcal{R}}_{t:t+K})$ that maps the simulator state $x_t$ and a future reference window to control actions under multi-body contact dynamics.
Crucially, the object's internal joints remain purely passive in simulation: no joint actuators are enabled and reference states $\widehat{q}_t$ are never prescribed during rollout, requiring all joint motion to arise from humanoid contact forces.

\subsection{Contact-Aware Articulated Reconstruction}
\label{sec:contact-aware-reconstruction}

Existing monocular articulated HOI methods typically fit part states frame by frame via propagation from a canonical frame~\citep{wang2026arthoi,huang2026arthoi}. Since past estimates are never revised, per-frame depth noise causes severe articulation jitter and hand-object contact drift, yielding kinematic references unsuitable for passive physics control.
To overcome this, we formulate articulated reconstruction as a global sequence optimization followed by contact-aware arm-hand alignment.
Specifically, we first extract per-frame metric depth $\{D_t\}_{t=1}^{T}$, part segmentation masks, dense 2D point tracks, and initial SMPL-X motion~\citep{yang2026sam3dbody} from $V$. 
In parallel, Qwen3.7-Plus~\citep{qwen2026qwen37plus} annotates the active part $\ell$ and its candidate contact region on $\mathcal{A}$, while Qwen3.8-Flash-Next~\citep{qwen2026qwen38flashnext} identifies the contact frames $\mathcal{C}$ and participating finger segments $\mathcal{F}_t$ in the video (\cref{app:implementation}).

\vspace{0.5ex}
\noindent\textbf{Trajectory-Level Articulation Tracking.}
To track the active part $\ell$ coherently over time, we sample $N$ 2D point tracks within its segmentation mask and back-project them using the metric depth maps $D_t$ to obtain 3D trajectories $\{X_{t,i} \in \mathbb{R}^3\}_{(t,i)\in\mathcal{V}}$.
Here, $\mathcal{V} \subset \{1,\dots,T\} \times \{1,\dots,N\}$ indexes the visible frame-track pairs.
Each track $i$ is anchored to a fixed surface point $\Phi_{\ell}(b_i) \in \mathbb{R}^3$ on the canonical CAD model of part $\ell$, parameterized by time-invariant barycentric coordinates $b_i$.
Given the estimated object scale $s$, root pose $T^o_t \in \mathrm{SE}(3)$, and joint state $q_t \in \mathbb{R}^J$, forward kinematics $\mathcal{T}_{\mathrm{cam}}(\cdot)$ maps the canonical surface point $\Phi_{\ell}(b_i)$ to the camera frame as $\mathcal{T}_{\mathrm{cam}}(T^o_t, s, q_t, \ell)\,\Phi_{\ell}(b_i) \in \mathbb{R}^3$.
Holding $T^o_t$ and $s$ fixed at their initial estimates, 
we optimize the full-sequence joint trajectory $q_{1:T}$ by minimizing:
\begin{multline}
\mathcal{L}_{\mathrm{traj}}(q_{1:T})
= \sum_{(t,i)\in\mathcal{V}} \bigl\|X_{t,i}-\mathcal{T}_{\mathrm{cam}}(T^o_t,s,q_t,\ell)\,\Phi_{\ell}(b_i)\bigr\|_1 \\
+ \lambda_{\mathrm{acc}}\sum_{t=2}^{T-1}\bigl\|q_{t+1}-2q_t+q_{t-1}\bigr\|_2^2
+ \lambda_{\mathrm{init}}\sum_{t<t_0}\bigl\|q_t-q^{\mathrm{init}}\bigr\|_2^2,
\label{eq:active-part-track}
\end{multline}
where $q^{\mathrm{init}}$ is the initial joint state, $t_0 = \min \mathcal{C}$ is the contact onset frame detected by the VLM, and $\lambda_{\mathrm{acc}}, \lambda_{\mathrm{init}}$ weight the acceleration smoothness and pre-contact clamping priors.
To avoid local minima, we initialize $q_{1:T}$ globally via the Viterbi algorithm~\citep{viterbi1967error} over discretized joint states by matching rendered and segmented part silhouettes before minimizing \cref{eq:active-part-track}.

\vspace{0.5ex}
\noindent\textbf{Contact-Aware HOI Alignment.}
We then fix the recovered articulation $q_{1:T}$ and align the human and object in two stages, as shown in \cref{fig:overview}.
\emph{(i) Object refinement.}
For real monocular videos, we optimize a sequence-wide correction to the object root pose and scale while holding the human motion and recovered articulation fixed.
For asset-conditioned generated videos, the camera, object root pose, and scale are known from the 3D scene layout and require no further refinement.
\emph{(ii) Contact-anchored arm--hand IK.}
With object placement fixed, we optimize the arm and finger joints to bring the VLM-identified finger regions onto the annotated contact region carried by the moving link.
To retain natural body motion and obtain smooth corrections over time, we parameterize these rotations with a residual cubic B-spline, $\theta_t=\bar{\theta}_t+\sum_k B_k(t)c_k$, where $\bar{\theta}_t$ is the initial pose, $B_k(t)$ are spline bases, and $c_k$ are sparse control points.
The human root and remaining joints retain their original motion.
We optimize the control points with the following objective:
\begin{equation}
\mathcal{E}_{\mathrm{align}}
= \lambda_c\mathcal{L}_{\mathrm{contact}}
+ \lambda_p\mathcal{L}_{\mathrm{pen}}
+ \sum_{t\in\mathcal{C}} \mathcal{E}_{\mathrm{grasp}}(t)
+ \mathcal{L}_{\mathrm{reg}}.
\label{eq:recon-objective}
\end{equation}
The contact and penetration losses place the manipulating hand on the moving contact surface while limiting intersections. 
We further integrate the grasp prior $\mathcal{E}_{\mathrm{grasp}}$ adapted from HOI-FHLI~\citep{wu2024hoifhli}, which regularizes the finger configuration toward plausible grasp poses and encourages force closure.
By refining the hand shape, this prior complements arm alignment and directly benefits physical execution, as confirmed by our ablation in \cref{tab:recon-to-physics}.
While the contact, penetration, and grasp terms act only during contact frames $\mathcal{C}$, the smoothness regularizer $\mathcal{L}_{\mathrm{reg}}$ penalizes joint velocity and acceleration across the entire sequence (\cref{app:implementation}).

\subsection{Articulation-Guided Physics Refinement}
\label{sec:articulation-guided-refiner}

While $\widehat{\mathcal{R}}_{1:T}$ specifies kinematic hand-link association, it lacks the contact forces needed to drive the passive joint in physics simulation.
Our proposed \textbf{Articulation-Guided Physics Refinement} (\cref{fig:overview}(b)) trains a closed-loop policy $\pi_{\theta}$ under contact dynamics by pairing a residual policy atop a frozen dexterous tracker with articulation-guided contact and progress rewards.

\vspace{0.5ex}
\noindent\textbf{Residual Policy Learning.}
To avoid learning whole-body balance and tracking from scratch within a fixed training budget, we build a closed-loop residual policy $\pi_{\theta}$ atop the frozen pretrained PHC-X tracker with articulated fingers~\citep{luo2024omnigrasp}.
At each step $t$, the frozen tracker outputs nominal PD targets $u^{\mathrm{ref}}_t$, and $\pi_{\theta}$ adds bounded corrections $\delta_t \in [-1, 1]$ across all body and finger degrees of freedom:
\begin{equation}
u^{\mathrm{pd}}_t = u^{\mathrm{ref}}_t + \sigma\,\delta_t,
\label{eq:residual-action}
\end{equation}
where $\sigma$ scales the residual.
With the frozen tracker maintaining base balance and coordinated whole-body motion, $\pi_{\theta}$ is trained with PPO~\citep{schulman2017ppo} using proprioception, reference states, fingertip contact forces, and relative hand-to-link poses as inputs.
This residual structure lets exploration focus on sustaining contact and actuating the passive joint, allowing each per-asset policy to train jointly across multiple interaction trajectories of that object.

\vspace{0.5ex}
\noindent\textbf{Articulation-Guided Rewards.}
Driving articulated parts requires placing fingers precisely on the actionable surface and applying sustained force along the constrained joint axis.
We formulate the total reward by combining a multiplicative tracking product with an additive actuation progress term:
\begin{equation}
r_t
= r^{\mathrm{body}}_t\, r^{\mathrm{obj}}_t\, r^{\mathrm{contact}}_t
+ w_{\mathrm{prog}}\, r^{\mathrm{prog}}_t,
\label{eq:reward}
\end{equation}
where $w_{\mathrm{prog}}$ weights joint progress, and $r^{\mathrm{body}}_t, r^{\mathrm{obj}}_t \in (0,1]$ are exponential tracking rewards for the human and object states (\cref{app:physics}).
To guide precise finger placement over active contact frames $t\in\mathcal{C}$, we define:
\begin{equation}
r^{\mathrm{contact}}_t = \frac{1}{|\mathcal{F}_t|}\sum_{j\in\mathcal{F}_t}\exp(-w_{\mathrm{c}}\,d_{t,j}),
\label{eq:contact-reward}
\end{equation}
averaging the exponential distance kernel over distances $d_{t,j}$ from each participating finger segment $j\in\mathcal{F}_t$ to its actionable link region (with $r^{\mathrm{contact}}_t = 1$ for $t \notin \mathcal{C}$).

However, relying solely on frame-wise tracking often fails to open or move the passive joint early in training, as any lag behind the advancing reference $\widehat{q}_t$ rapidly decays the instantaneous tracking product and leaves little gradient for exploration.
We therefore add a phase-grounded progress reward $r^{\mathrm{prog}}_t \in [0, 1]$ outside the product that directly rewards advancing the joint from its initial state, giving immediate positive feedback once the link moves so the policy quickly learns to actuate the joint.
During each active opening or closing phase with direction $d\in\{-1,1\}$, reference amplitude $\Delta$, and phase-entry joint state $q_e$ ($r^{\mathrm{prog}}_t = 0$ on stationary phases), we define normalized progress $\bar p_t = \operatorname{clip}\!\big(d(q_t-q_e)/\Delta,\, 0,\, \operatorname{clip}(d(\widehat{q}_t-q_e)/\Delta,0,1)\big)$ and set:
\begin{equation}
r^{\mathrm{prog}}_t = \left[m_t\,\bar p_t
- \lambda_{\mathrm{o}}\,\left[\frac{d\,(q_t-\widehat{q}_t)}{\Delta}\right]_+\right]_+,
\label{eq:progress}
\end{equation}
where $[\cdot]_+=\max(\cdot,0)$, $\lambda_{\mathrm{o}}$ penalizes overshoot beyond $\widehat{q}_t$, and $m_t \in [0, 1]$ is the fraction of target finger segments in active contact with the link.
Weighting $\bar p_t$ by $m_t$ credits joint motion only under active finger contact, whereas bounding $\bar p_t$ by the reference progress and penalizing overshoot keeps actuation paced with $\widehat{q}_t$.

\section{Experiments}
\label{sec:experiments}

\begin{table*}[t]
  \caption{\textbf{Articulated HOI reconstruction on videos generated with RoboCasa~\citep{nasiriany2024robocasa} assets.} All methods are evaluated on the same 203 videos.
  We report the average optimization time per video on a single NVIDIA V100 GPU, excluding shared preprocessing. $^\dagger$ denotes adaptation to the matched-CAD setting. Best and second-best results are shown in \textbf{bold} and \underline{underlined}, respectively.}
  \vspace{-0.4em}

  \label{tab:generated-video}
  \centering
  \small
  \setlength{\tabcolsep}{4pt}

  \resizebox{0.95\textwidth}{!}{%
  \renewcommand{\arraystretch}{1.08}
  \begin{tabular}{lcccccccc}
    \toprule
    & \multicolumn{1}{c}{\textbf{Efficiency}} & \multicolumn{2}{c}{\textbf{Interaction}} & \multicolumn{3}{c}{\textbf{Input consistency}} & \multicolumn{2}{c}{\textbf{Smoothness} ($\times 10^{-2}$)} \\
    \cmidrule(lr){2-2}\cmidrule(lr){3-4}\cmidrule(lr){5-7}\cmidrule(l){8-9}
    \textbf{Method} & Time (min)$\,\downarrow$ & Contact (cm)$\,\downarrow$ & Pen. (\%)$\,\downarrow$
    & Part IoU$\,\uparrow$ & Body 2D$\,\downarrow$ & Hand 2D$\,\downarrow$
    & Human$\,\downarrow$ & Object$\,\downarrow$ \\
    \midrule
    {ArtHOI-Hand$^\dagger$~\citep{wang2026arthoi}} & 38.98 & 45.59 & 11.18 & 0.61 & -- & 26.48 & 1.63 & \underline{0.23} \\
    ArtHOI-GS$^\dagger$~\citep{huang2026arthoi} & 40.38 & \underline{21.15} & \underline{2.31} & \textbf{0.65} & 11.78 & 15.94 & 2.04 & 0.29 \\
    \midrule
    Ours w/o HOI Align & \textbf{1.06} & 26.85 & \textbf{2.23} & \underline{0.63} & \textbf{5.03} & \textbf{6.32} & \textbf{0.25} & \textbf{0.14} \\
    \rowcolor{OursBlue}
    \textbf{Ours} & \underline{4.23} & \textbf{4.69} & 2.63 & \underline{0.63} & \underline{6.92} & \underline{14.84} & \underline{0.29} & \textbf{0.14} \\
    \bottomrule
  \end{tabular}%
  }

\end{table*}
\vspace{0.8em}

\begin{table*}[t]
  \caption{\textbf{Articulated HOI reconstruction on real D3D-HOI~\citep{xu2021d3dhoi} videos with corresponding CAD models.} $^\dagger$ denotes adaptation to the matched-CAD setting. Best and second-best results are shown in \textbf{bold} and \underline{underlined}, respectively.}
  \vspace{-0.4em}

  \label{tab:reconstruction}
  \centering
  \setlength{\tabcolsep}{7pt}
  \small
  \resizebox{0.95\textwidth}{!}{%
  \renewcommand{\arraystretch}{1.08}
  \begin{tabular}{lcccccc}
    \toprule
    & \multicolumn{2}{c}{\textbf{Interaction}} & \multicolumn{2}{c}{\textbf{Articulation}}
    & \multicolumn{2}{c}{\textbf{Object Pose}} \\
    \cmidrule(lr){2-3}\cmidrule(lr){4-5}\cmidrule(l){6-7}
    \textbf{Method} & Contact (cm)$\,\downarrow$ & Pen. (\%)$\,\downarrow$
    & Revolute (deg)$\,\downarrow$ & Prismatic (cm)$\,\downarrow$
    & Orient. (deg)$\,\downarrow$ & Loc. (cm)$\,\downarrow$ \\
    \midrule
    D3D-HOI~\citep{xu2021d3dhoi} & {37.48} & \textbf{{3.79}} & \underline{{16.13}} & {3.40} & {18.29} & \underline{{31.17}} \\
    ArtHOI-Hand$^\dagger$~\citep{wang2026arthoi} & {24.16} & {15.49} & {25.96} & {8.09} & \underline{{10.17}} & {32.16} \\
    ArtHOI-GS$^\dagger$~\citep{huang2026arthoi} & {19.93} & {4.24} & {19.38} & \textbf{{2.70}} & \textbf{{10.15}} & {32.25} \\
    \midrule
    Ours w/o HOI Align & \underline{{18.38}} & \underline{{3.87}} & \textbf{{12.54}} & {3.12} & \underline{{10.17}} & {32.16} \\
    \rowcolor{OursBlue}
    \textbf{Ours} & \textbf{{7.79}} & {4.67} & \textbf{{12.54}} & \underline{{3.00}} & {11.50} & \textbf{{30.98}} \\
    \bottomrule
  \end{tabular}%
  }

\end{table*}

We evaluate \om{} across three complementary sources of articulated interaction priors to answer four questions:
(1) Does Contact-Aware Articulated Reconstruction recover accurate object articulation and physically plausible hand-link contact from both synthetic and real monocular videos (\cref{sec:reconstruction-experiments})?
(2) Can Articulation-Guided Physics Refinement convert imperfect kinematic references into full-body control that actuates passive joints through contact, efficiently enough for one policy to cover many trajectories of an asset (\cref{sec:physics-experiments})?
(3) How do individual reconstruction constraints and physics learning components contribute to execution performance (\cref{sec:ablation-studies})?
(4) Do the physically refined interactions support downstream object-goal policy learning and cross-embodiment transfer to a humanoid robot (\cref{sec:applications})?

\subsection{Experimental Setup}
\label{sec:experimental-setup}

\paragraph{Benchmarks.}
For \emph{generated videos}, we follow GRAIL~\citep{xie2026grail} to synthesize 276 clips conditioned on RoboCasa assets~\citep{nasiriany2024robocasa}, retaining 203 clips after artifact filtering for reconstruction and 128 trajectories across 16 CAD assets for per-CAD physics evaluation.
For \emph{real videos}, we use D3D-HOI~\citep{xu2021d3dhoi}, which provides a matched PartNet-Mobility~\citep{xiang2020sapien} CAD model for each clip while leaving camera, object pose, scale, and articulation to be estimated; after filtering non-hand and negligible-motion clips, we evaluate 253 videos for reconstruction and 251 trajectories across 16 assets for physics.
For \emph{motion capture}, we use 178 sequences across 8 assets from ARCTIC~\citep{fan2023arctic} to evaluate physics control without reconstruction error.
Dataset details are provided in \cref{app:experimental-details}.

\paragraph{Baselines.}
For reconstruction, we compare \om{} against matched-CAD adaptations of ArtHOI-GS~\citep{huang2026arthoi} and ArtHOI-Hand~\citep{wang2026arthoi}, providing all three methods with the same CAD model, camera, masks, and initial object state.
ArtHOI-GS and \om{} additionally share the same initial full-body pose estimates, and both baseline adaptations recover part motion by fitting the active joint of the CAD model.
On D3D-HOI (\cref{tab:reconstruction}), we also include the object predictions released with the dataset.
For physics execution, we compare with InterMimic~\citep{xu2025intermimic},
HDMI~\citep{weng2025hdmi}, and CoDA~\citep{pi2025coda}.
All physics methods execute the same frozen references under a shared training budget of 4096 parallel environments and 1000 policy updates, and are evaluated under the same protocol.
Implementation details of the adapted baselines are provided in \cref{app:baseline-implementation}.

\paragraph{Metrics.}
Since generated videos lack 3D ground truth, we evaluate reconstruction quality by hand-link contact and interpenetration, 2D consistency with the input video, and temporal smoothness.
On D3D-HOI, we additionally measure joint-state and object pose errors against ground-truth annotations.
For physics execution, we report the success rate, global human pose error (G-MPJPE), active-frame articulation error (AAE), normalized joint progress, and contact coverage of the moving link.
Formal definitions of all metrics are given in \cref{app:physics-evaluation-protocol,app:recon-interaction-metrics}.

\subsection{Articulated HOI Reconstruction}
\label{sec:reconstruction-experiments}

\begin{figure*}[t]
  \centering
  \includegraphics[width=0.9\textwidth]{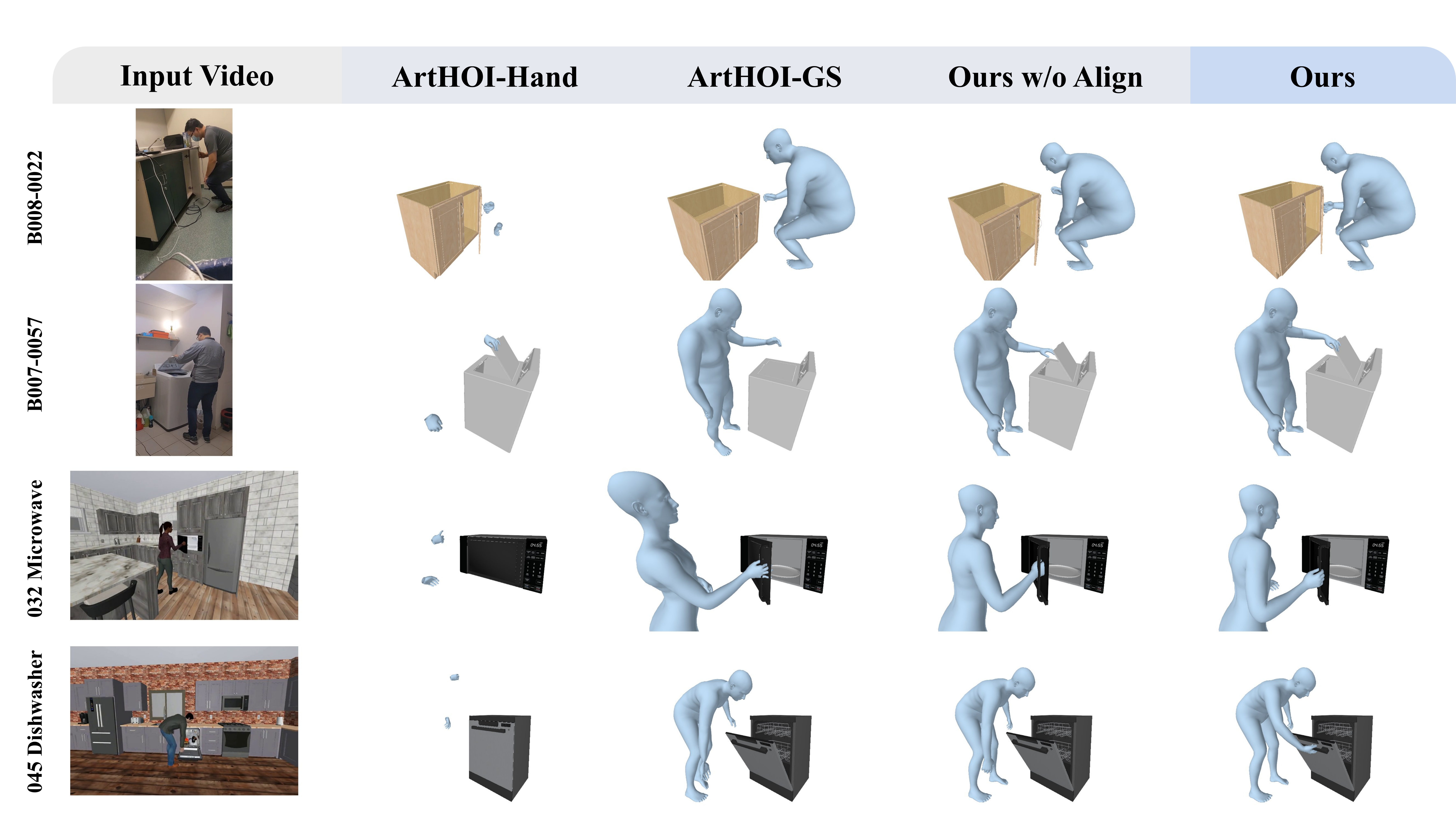}
  \vspace{-0.5em}
  \caption{\textbf{Qualitative comparison of articulated HOI reconstruction.}
  On real D3D-HOI (top) and generated RoboCasa (bottom) videos, the baselines misestimate the articulation and leave the hand away from the handle, whereas \om{} recovers accurate link motion and places the hand on the moving handle, providing contact-consistent references for physics execution.}
  \label{fig:qual-recon}
\end{figure*}

As reported in \cref{tab:generated-video,tab:reconstruction}, \om{} achieves a markedly lower hand-link contact distance than all baselines on both generated and real videos, while optimizing nearly $10\times$ faster than the ArtHOI baselines on generated videos.
ArtHOI-Hand and ArtHOI-GS fit the part motion frame by frame.
On generated videos, ArtHOI-GS attains the highest part-mask IoU, with larger 3D hand-link distances and less smooth motion than \om{}.
\om{} also attains the lowest revolute joint error on D3D-HOI.
Notably, HOI alignment accounts for most of the contact improvement, and \cref{fig:qual-recon} demonstrates that it keeps the hand on the moving handle throughout the articulation.

\subsection{Physics Execution}
\label{sec:physics-experiments}

We next execute the reconstructed and captured references in simulation, training one policy per CAD asset on all reference trajectories of that asset, as reported in \cref{tab:physics-main}.

\begin{table*}[t]
  \caption{\textbf{Physics execution with per-CAD policies.}
  Each policy is trained on all reference trajectories of one articulated asset.
  Within each dataset, all methods use the same frozen references, training budget, and evaluation protocol.
  Success rate (SR) uses the phase-duration-weighted mean progress criterion $P \geq 0.5$ (\cref{app:metric-audit}).
  Results are averaged over three random seeds.
  Best results are in bold. $^\dagger$ denotes adaptation to the articulated-object setting.}
  \vspace{-0.4em}
  \label{tab:physics-main}

  \centering
  \small
  \setlength{\tabcolsep}{6pt}
  \renewcommand{\arraystretch}{1.08}
  \setlength{\aboverulesep}{1pt}
  \setlength{\belowrulesep}{1pt}
  \resizebox{0.88\textwidth}{!}{%
  \begin{tabularx}{\textwidth}{lcc*{4}{>{\centering\arraybackslash}X}}
    \toprule
    \multirow{2}{*}{\textbf{Method}}
      & \multirow{2}{*}{\textbf{SR}~(\%)$\,\uparrow$}
      & \multirow{2}{*}{\textbf{G-MPJPE}~(mm)$\,\downarrow$}
      & \multicolumn{2}{c}{\textbf{AAE}$\,\downarrow$}
      & \multirow{2}{*}{\textbf{Prog.} (\%)$\,\uparrow$}
      & \multirow{2}{*}{\textbf{Cov.} (\%)$\,\uparrow$} \\
    \cmidrule(lr){4-5}
      & & & Rev. ($^\circ$) & Pris. (cm) & & \\
    \midrule
    \PhysicsBenchRow{RoboCasa}{Generated Video}{GenBg}{GenFg}{16}{128}
    {InterMimic$^\dagger$~\citep{xu2025intermimic}} & {2.3} & {246.3} & {32.50} & {15.98} & {7.4} & {29.9} \\
    {HDMI~\citep{weng2025hdmi}} & {4.7} & {355.7} & {26.54} & {15.44} & {10.4} & {28.2} \\
    {CoDA~\citep{pi2025coda}} & {25.8} & {160.6} & {23.97} & {\textbf{4.28}} & {29.1} & {43.6} \\
        \rowcolor{OursBlue}
    {\textbf{Ours}} & {\textbf{53.1}} & {\textbf{100.6}} & {\textbf{17.17}} & {5.27} & {\textbf{55.9}} & {\textbf{81.7}} \\
    \midrule
    \PhysicsBenchRow{D3D-HOI}{Real Video}{RealBg}{RealFg}{16}{251}
    {InterMimic$^\dagger$~\citep{xu2025intermimic}} & {17.5} & {270.9} & {25.94} & {12.52} & {33.8} & {35.6} \\
    {HDMI~\citep{weng2025hdmi}} & {4.4} & {323.5} & {19.25} & {12.37} & {18.7} & {22.3} \\
    {CoDA~\citep{pi2025coda}} & {38.6} & {235.3} & {16.02} & {10.03} & {49.6} & {39.8} \\
    \rowcolor{OursBlue}
    {\textbf{Ours}} & {\textbf{68.1}} & {\textbf{115.4}} & {\textbf{11.66}} & {\textbf{7.75}} & {\textbf{73.9}} & {\textbf{78.3}} \\
    \midrule
    \PhysicsBenchRow{ARCTIC}{Captured Reference}{CaptureBg}{CaptureFg}{8}{178}
    {InterMimic$^\dagger$~\citep{xu2025intermimic}} & {27.5} & {75.5} & {56.60} & {--} & {46.6} & {\textbf{83.9}} \\
    {HDMI~\citep{weng2025hdmi}} & {10.7} & {228.6} & {55.11} & {--} & {41.8} & {35.7} \\
    {CoDA~\citep{pi2025coda}} & {19.1} & {172.7} & {56.05} & {--} & {38.6} & {17.7} \\
        \rowcolor{OursBlue}
    {\textbf{Ours}} & {\textbf{51.1}} & {\textbf{71.5}} & {\textbf{43.31}} & {--} & {\textbf{61.6}} & {72.5} \\
    \bottomrule
  \end{tabularx}
  }
\end{table*}

\paragraph{Execution from video references.}
With one policy per CAD asset, \om{} achieves the highest success, joint progress, and contact coverage on both real and generated videos, improving success over the strongest baseline from 38.6\% to 68.1\% on D3D-HOI.
On the RoboCasa physics subset, more than half of the evaluated references are executed successfully, showing that generated videos can provide physically usable interaction priors.
Building on a pretrained whole-body tracker lets \om{} focus learning on contact and actuation, whereas the baselines learn balance, tracking, and contact jointly within a single policy.
In the 45-case D3D-HOI per-trajectory evaluation, \om{} still achieves the best body tracking, articulation accuracy, joint progress, and contact coverage (\cref{app:additional-results}).

\begin{figure*}[t]
  \centering
  \includegraphics[width=0.9\textwidth]{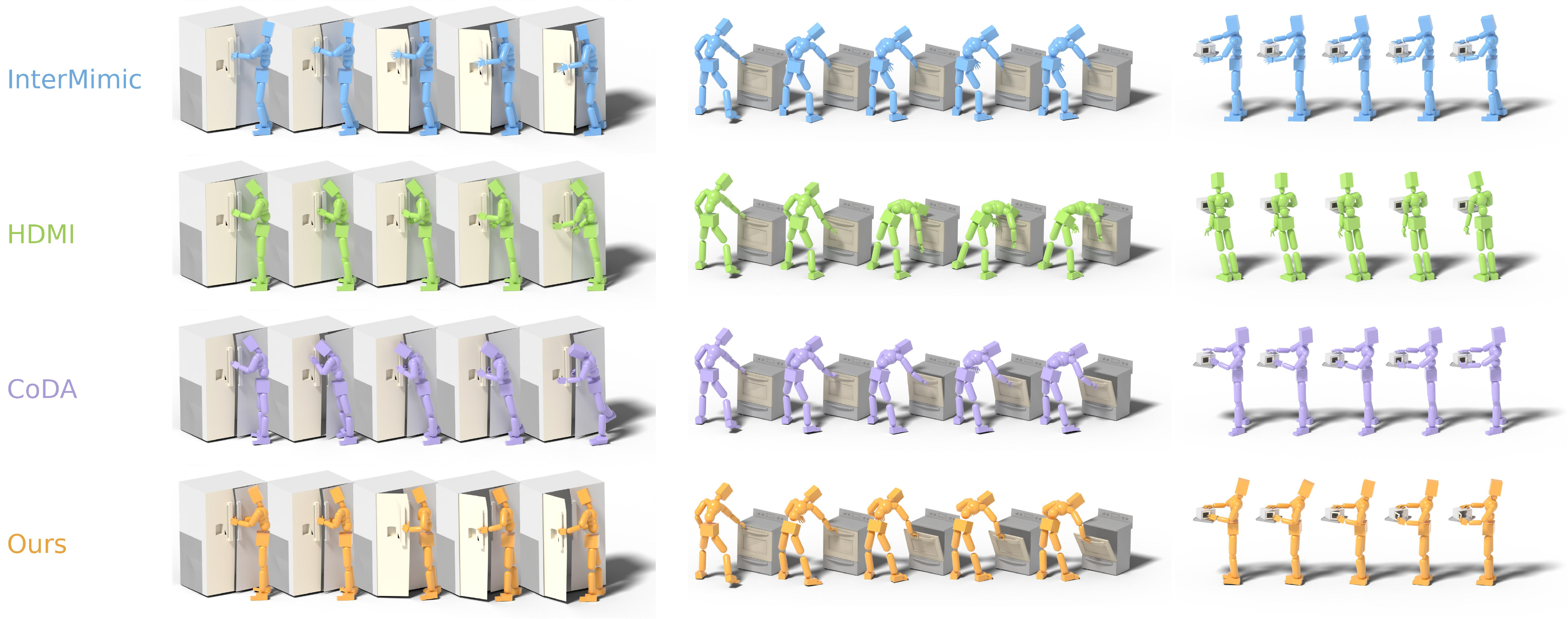}
  \vspace{-1.0em}
  \caption{\textbf{Qualitative comparison of physics execution.}
  All methods follow the same reference for a refrigerator (left), an oven (middle), and a microwave (right), shown at five shared timesteps.
  \om{} keeps the hand on the moving part and advances each joint along the reference, whereas the baselines stall the joint or lose hand contact.}
  \vspace{-0.4em}
  \label{fig:qual-physics}
\end{figure*}

\paragraph{Dexterous execution from captured references.}
ARCTIC poses a different challenge, as its small tabletop objects require dexterous finger manipulation, while its motion-captured references remove reconstruction error so that performance reflects control alone.
Here, \om{} turns contact into actuation most effectively, lifting success over the strongest baseline from 27.5\% to 51.1\% with the lowest tracking error.
InterMimic attains the highest contact coverage yet advances the joint far less.
\cref{fig:qual-physics} shows representative rollouts, in which only \om{} sustains hand contact while advancing the passive joint along the reference.

\subsection{Ablation Studies}
\label{sec:ablation-studies}

\noindent\textbf{Effect of contact-aware reconstruction on physical execution.}
With the physics refiner fixed, references from \om{} yield the best success, tracking accuracy, joint progress, and contact coverage in \cref{tab:recon-to-physics}, improving success from 40.6\% with ArtHOI-GS to 53.1\% with lower motion jerk.
Among our reconstruction components, HOI alignment contributes most to success, joint progress, and especially contact coverage, as aligning the hand with the moving link helps the policy establish contact.
The grasp prior contributes to tracking accuracy and also raises success by refining finger placement on the handle.
\begin{table}[t]
  \caption{\textbf{Effect of articulated HOI reconstruction on physics execution.} The same physics refiner executes each reconstruction on 128 trajectories from 16 RoboCasa~\citep{nasiriany2024robocasa} assets.
  }
  \vspace{-0.4em}
  \label{tab:recon-to-physics}
  \centering
  \setlength{\tabcolsep}{6pt}
  \resizebox{\columnwidth}{!}{%
  \renewcommand{\arraystretch}{1.08}
  \begin{tabular}{lccccc}
    \toprule
    \multirow{2}{*}{\textbf{Reconstruction}} & \multicolumn{5}{c}{\textbf{Physics execution}} \\
    \cmidrule(l){2-6}
    & SR$\uparrow$ & G-MPJPE$\downarrow$ & Prog.$\uparrow$ & Cov.$\uparrow$ & Jerk$\downarrow$ \\
    \midrule
    ArtHOI-GS$^\dagger$~\citep{huang2026arthoi} & 40.6 & 155.8 & 48.4 & 66.7 & 2.20 \\
    Ours w/o HOI Align & 46.9 & 109.0 & 52.5 & 70.0 & 0.95 \\
    Ours w/o Grasp Prior & 50.8 & 112.0 & 53.7 & 80.1 & \textbf{0.92} \\
    \rowcolor{OursBlue}
    \textbf{Ours} & 
    \textbf{53.1} & \textbf{100.6} & \textbf{55.9} & \textbf{81.7} & 1.01 \\
    \bottomrule
  \end{tabular}
  }
\end{table}

\begin{table}[t]
  \caption{\textbf{Ablation of the physics refiner on D3D-HOI~\citep{xu2021d3dhoi}.} Each row adds one component to a vanilla policy, and all variants share the same references and training budget. 
  }
  \vspace{-0.4em}
  \label{tab:ablation-main}
  \centering
  \setlength{\tabcolsep}{6pt}
  \resizebox{\columnwidth}{!}{%
  \renewcommand{\arraystretch}{1.08}
  \begin{tabular}{ccccccc}
    \toprule
    \multicolumn{3}{c}{\textbf{Design choices}} & \multicolumn{4}{c}{\textbf{Physics execution}} \\
    \cmidrule(lr){1-3}\cmidrule(l){4-7}
    Residual & Contact & Progress & SR$\uparrow$ & G-MPJPE$\downarrow$ & Prog.$\uparrow$ & Cov.$\uparrow$ \\
    \midrule
      &  &  & 34.7 & 219.4 & 48.2 & 73.1 \\
    \checkmark &  &  & 62.9 & 112.3 & 66.0 & 75.1 \\
    \checkmark & \checkmark &  & 65.7 & \textbf{101.8} & 68.3 & \textbf{79.1} \\
    \rowcolor{OursBlue}
    \checkmark & \checkmark & \checkmark & \textbf{68.1} & 115.4 & \textbf{73.9} & 78.3 \\
    \bottomrule
  \end{tabular}
  }
\end{table}

\paragraph{Components of the physics refiner.}
\cref{tab:ablation-main} starts from a direct policy trained without the pretrained tracker and adds residual control, fine-grained contact, and the phase-grounded progress reward one at a time.
Residual control on top of the frozen whole-body tracker brings the largest gain, lifting success from 34.7\% to 62.9\% and roughly halving the tracking error.
Fine-grained contact rewards improve all four metrics, most notably tracking accuracy and contact coverage.
The progress reward, which encourages advancing the passive joint along the reference, yields the highest success and joint progress at a moderate tracking cost.

\subsection{Downstream Applications}
\label{sec:applications}

We further demonstrate two downstream applications of the physically executable trajectories produced by \om{}.
\emph{(1) Object-goal policy distillation.}
Following InterPrior~\citep{xu2026interprior}, we distill each full-reference physics teacher into a goal-conditioned student policy through online distillation followed by RL finetuning.
The student has no access to the full motion reference and observes only the current humanoid and object states with a target joint trajectory.
During training, we randomly perturb the initial humanoid position and heading by small offsets, whereas at test time we apply larger unseen perturbations of up to 20\,cm and 30$^\circ$ and add smooth perturbations bounded by 10$^\circ$ to the target joint sequence.
As shown in \cref{tab:student-goal}, the students retain over 30\% success under these perturbations, demonstrating robustness beyond the training distribution.
\emph{(2) Cross-embodiment transfer.}
We retarget our references to a Unitree G1~\citep{unitreeg1} with SUGAR~\citep{wu2026sugar} and adapt our articulation-aware rewards, enabling the robot to approach, grasp, and actuate articulated appliances in simulation.
Training details of the students and qualitative G1 demonstrations in simulation are provided in \cref{app:student-diagnostics,app:g1-application}.

\begin{table}[t]
  \caption{\textbf{Object-goal policy distillation.} Each student tracks a commanded joint trajectory without human reference motion. 
  }
  \label{tab:student-goal}
  \centering
  \scriptsize
  \setlength{\tabcolsep}{1.2pt}
  \renewcommand{\arraystretch}{1.08}
  \begin{tabular*}{\columnwidth}{@{\extracolsep{\fill}}l*{9}{c}@{}}
    \toprule
    \multirow{2}{*}{\textbf{Asset}}
      & \multicolumn{3}{c}{\shortstack{\textbf{No}\\\textbf{Perturb.}}}
      & \multicolumn{3}{c}{\shortstack{\textbf{Human}\\\textbf{Perturb.}}}
      & \multicolumn{3}{c}{\shortstack{\textbf{Human + Goal}\\\textbf{Perturb.}}} \\
    \cmidrule(lr){2-4}\cmidrule(lr){5-7}\cmidrule(lr){8-10}
      & SR $\uparrow$ & AAE $\downarrow$ & Fall $\downarrow$
      & SR $\uparrow$ & AAE $\downarrow$ & Fall $\downarrow$
      & SR $\uparrow$ & AAE $\downarrow$ & Fall $\downarrow$ \\
    \midrule
    Microwave & 69.4 & 5.42 & 8.3 & 34.4 & 11.32 & 23.1 & 31.7 & 11.63 & 23.3 \\
    Dishwasher & 90.9 & 2.09 & 0.0 & 42.7 & 12.77 & 22.7 & 43.6 & 12.54 & 23.6 \\
    Refrigerator & 71.4 & 0.69 & 0.0 & 45.7 & 1.55 & 14.3 & 40.0 & 1.51 & 15.7 \\
    \bottomrule
  \end{tabular*}
\end{table}

\section{Conclusion}
\label{sec:conclusion}

We presented \om{}, a scalable and efficient video-to-physics framework that turns monocular interaction videos into physically executable humanoid skills with articulated objects.
Its Contact-Aware Articulated Reconstruction recovers kinematic references with more accurate hand-link contact and smoother motion while optimizing nearly $10\times$ faster than frame-by-frame baselines on generated videos.
Articulation-Guided Physics Refinement leverages a residual policy on a frozen dexterous tracker with articulation-guided rewards to turn reconstructed trajectories into physically executable humanoid skills;
under the same training budget, it reaches higher success than physics-based HOI baselines on diverse interaction sources.
Better reconstructions in turn yield better execution, and preliminary results on policy distillation and humanoid transfer show potential for embodied learning.

\section*{Acknowledgments}
We sincerely thank Luyang Niu and Yue Chen for their insightful feedback and valuable discussions
throughout the course of this project.

\section*{LLM Usage Statement}
Generative AI tools and Large Language Models (including Codex) were used to assist with manuscript language polishing (improving wording, grammar, and structural clarity), preliminary literature discovery, code implementation, and scripting for data visualization. As part of the method, Qwen3.7-Plus~\citep{qwen2026qwen37plus} selects the target object and active part in RoboCasa scenes. Qwen3.8-Flash-Next~\citep{qwen2026qwen38flashnext} labels hand-object contact and screens generated videos. The MiniMax H3~\citep{minimax2026h3} image-to-video model generates the interaction videos used in our experiments. All research ideation, methodological design, and analysis were conducted by the authors. The authors have thoroughly reviewed and verified all AI-assisted outputs and take full responsibility for the integrity and content of this paper.
\vspace{3ex}

{
  \small
  \bibliographystyle{ieeenat_fullname}
  \bibliography{main}

@article{xu2021d3dhoi,
  title={{D3D-HOI}: Dynamic {3D} Human-Object Interactions from Videos},
  author={Xu, Xiang and Joo, Hanbyul and Mori, Greg and Savva, Manolis},
  journal={arXiv preprint arXiv:2108.08420},
  year={2021},
  eprint={2108.08420},
  archivePrefix={arXiv},
  url={https://arxiv.org/abs/2108.08420}
}

@inproceedings{todorov2012mujoco,
  title={{MuJoCo}: A Physics Engine for Model-Based Control},
  author={Todorov, Emanuel and Erez, Tom and Tassa, Yuval},
  booktitle={2012 IEEE/RSJ International Conference on Intelligent Robots and Systems},
  pages={5026--5033},
  year={2012},
  publisher={IEEE},
  doi={10.1109/IROS.2012.6386109},
  url={https://doi.org/10.1109/IROS.2012.6386109}
}

@inproceedings{xiang2020sapien,
  title={{SAPIEN}: A {SimulAted Part-Based Interactive ENvironment}},
  author={Xiang, Fanbo and Qin, Yuzhe and Mo, Kaichun and Xia, Yikuan and Zhu, Hao and Liu, Fangchen and Liu, Minghua and Jiang, Hanxiao and Yuan, Yifu and Wang, He and Yi, Li and Chang, Angel X. and Guibas, Leonidas J. and Su, Hao},
  booktitle={Proceedings of the IEEE/CVF Conference on Computer Vision and Pattern Recognition},
  pages={11097--11107},
  year={2020},
  url={https://openaccess.thecvf.com/content_CVPR_2020/html/Xiang_SAPIEN_A_SimulAted_Part-Based_Interactive_ENvironment_CVPR_2020_paper.html}
}

@misc{unitreeg1,
  title={Unitree G1 Humanoid},
  author={{Unitree Robotics}},
  year={2024},
  url={https://www.unitree.com/g1/}
}

@inproceedings{qian2022understanding,
  title={Understanding {3D} Object Articulation in Internet Videos},
  author={Qian, Shengyi and Jin, Linyi and Rockwell, Chris and Chen, Siyi and Fouhey, David F.},
  booktitle={Proceedings of the IEEE/CVF Conference on Computer Vision and Pattern Recognition},
  pages={1599--1609},
  year={2022}
}

@inproceedings{fan2023arctic,
  title={{ARCTIC}: A Dataset for Dexterous Bimanual Hand-Object Manipulation},
  author={Fan, Zicong and Taheri, Omid and Tzionas, Dimitrios and Kocabas, Muhammed and Kaufmann, Manuel and Black, Michael J. and Hilliges, Otmar},
  booktitle={Proceedings of the IEEE/CVF Conference on Computer Vision and Pattern Recognition},
  pages={12943--12954},
  year={2023},
  url={https://openaccess.thecvf.com/content/CVPR2023/html/Fan_ARCTIC_A_Dataset_for_Dexterous_Bimanual_Hand-Object_Manipulation_CVPR_2023_paper.html}
}

@inproceedings{bansal2026hopformer,
  title={Towards In-the-Wild Egocentric {3D} Hand-Object Pose Estimation},
  author={Bansal, Siddhant and Zhu, Zhifan and Tripathi, Shashank and Zhao, Jiahe and Black, Michael J. and Damen, Dima},
  booktitle={European Conference on Computer Vision},
  year={2026}
}

@inproceedings{pi2025coda,
  title={{CoDA}: Coordinated Diffusion Noise Optimization for Whole-Body Manipulation of Articulated Objects},
  author={Pi, Huaijin and Cen, Zhi and Dou, Zhiyang and Komura, Taku},
  booktitle={Advances in Neural Information Processing Systems},
  volume={38},
  pages={31758--31780},
  year={2025},
  doi={10.52202/085713-1067}
}

@article{weng2025hdmi,
  title={{HDMI}: Learning Interactive Humanoid Whole-Body Control from Human Videos},
  author={Weng, Haoyang and Li, Yitang and Sobanbabu, Nikhil and Wang, Zihan and Luo, Zhengyi and He, Tairan and Ramanan, Deva and Shi, Guanya},
  journal={arXiv preprint arXiv:2509.16757},
  year={2025},
  eprint={2509.16757},
  archivePrefix={arXiv},
  url={https://arxiv.org/abs/2509.16757}
}

@article{xie2026grail,
  title={{GRAIL}: Generating Humanoid Loco-Manipulation from World Models},
  author={Xie, Tianyi and Zhang, Haotian and Park, Jinhyung and Wang, Zi and Wen, Bowen and Li, Jiefeng and Li, Xueting and Ben, Qingwei and Weng, Haoyang and Ye, Yufei and Minor, David and Wang, Tingwu and Jiang, Chenfanfu and Fidler, Sanja and Kautz, Jan and Fan, Linxi and Zhu, Yuke and Luo, Zhengyi and Iqbal, Umar and Yuan, Ye},
  journal={arXiv preprint arXiv:2606.05160},
  year={2026},
  eprint={2606.05160},
  archivePrefix={arXiv},
  url={https://arxiv.org/abs/2606.05160}
}

@inproceedings{xu2025intermimic,
  title={{InterMimic}: Towards Universal Whole-Body Control for Physics-Based Human-Object Interactions},
  author={Xu, Sirui and Ling, Hung Yu and Wang, Yu-Xiong and Gui, Liang-Yan},
  booktitle={Proceedings of the IEEE/CVF Conference on Computer Vision and Pattern Recognition},
  pages={12266--12277},
  year={2025},
  doi={10.1109/CVPR52734.2025.01145}
}

@inproceedings{xu2026interprior,
  title={{InterPrior}: Scaling Generative Control for Physics-Based Human-Object Interactions},
  author={Xu, Sirui and Schulter, Samuel and Ziyadi, Morteza and He, Xialin and Fei, Xiaohan and Wang, Yu-Xiong and Gui, Liangyan},
  booktitle={Proceedings of the IEEE/CVF Conference on Computer Vision and Pattern Recognition},
  year={2026},
  eprint={2602.06035},
  archivePrefix={arXiv},
  url={https://arxiv.org/abs/2602.06035}
}

@inproceedings{huang2026repho,
  title={Recovering Physically Plausible Human-Object Interactions from Monocular Videos},
  author={Huang, Dingbang and Vouga, Etienne and Huang, Qixing and Pavlakos, Georgios},
  booktitle={Proceedings of the IEEE/CVF Conference on Computer Vision and Pattern Recognition},
  year={2026},
  eprint={2606.05359},
  archivePrefix={arXiv},
  url={https://arxiv.org/abs/2606.05359}
}

@article{kim2026devi,
  title={{DeVI}: Physics-based Dexterous Human-Object Interaction via Synthetic Video Imitation},
  author={Kim, Hyeonwoo and Kim, Jeonghwan and Cho, Kyungwon and Joo, Hanbyul},
  journal={arXiv preprint arXiv:2604.20841},
  year={2026},
  eprint={2604.20841},
  archivePrefix={arXiv},
  primaryClass={cs.CV},
  url={https://arxiv.org/abs/2604.20841}
}

@article{wang2023physhoi,
  title={{PhysHOI}: Physics-Based Imitation of Dynamic Human-Object Interaction},
  author={Wang, Yinhuai and Lin, Jing and Zeng, Ailing and Luo, Zhengyi and Zhang, Jian and Zhang, Lei},
  journal={arXiv preprint arXiv:2312.04393},
  year={2023},
  eprint={2312.04393},
  archivePrefix={arXiv},
  url={https://arxiv.org/abs/2312.04393}
}

@inproceedings{kim2025parahome,
  title={{ParaHome}: Parameterizing Everyday Home Activities Towards {3D} Generative Modeling of Human-Object Interactions},
  author={Kim, Jeonghwan and Kim, Jisoo and Na, Jeonghyeon and Joo, Hanbyul},
  booktitle={Proceedings of the IEEE/CVF Conference on Computer Vision and Pattern Recognition},
  year={2025}
}

@inproceedings{wang2026arthoi,
  title={{ArtHOI}: Taming Foundation Models for Monocular {4D} Reconstruction of Hand-Articulated-Object Interactions},
  author={Wang, Zikai and Zhang, Zhilu and Wang, Yiqing and Li, Hui and Zuo, Wangmeng},
  booktitle={Proceedings of the IEEE/CVF Conference on Computer Vision and Pattern Recognition},
  pages={15998--16009},
  year={2026},
  eprint={2603.25791},
  archivePrefix={arXiv},
  url={https://openaccess.thecvf.com/content/CVPR2026/html/Wang_ArtHOI_Taming_Foundation_Models_for_Monocular_4D_Reconstruction_of_Hand-Articulated-Object_CVPR_2026_paper.html}
}

@article{huang2026arthoi,
  title={{ArtHOI}: Articulated Human-Object Interaction Synthesis by {4D} Reconstruction from Video Priors},
  author={Huang, Zihao and Liu, Tianqi and Chen, Zhaoxi and Xu, Shaocong and Zhang, Saining and Xiao, Lixing and Cao, Zhiguo and Li, Wei and Zhao, Hao and Liu, Ziwei},
  journal={arXiv preprint arXiv:2603.04338},
  year={2026},
  eprint={2603.04338},
  archivePrefix={arXiv},
  url={https://arxiv.org/abs/2603.04338}
}

@article{liu2025videoartgs,
  title={{VideoArtGS}: Building Digital Twins of Articulated Objects from Monocular Video},
  author={Liu, Yu and Jia, Baoxiong and Lu, Ruijie and Gan, Chuyue and Chen, Huayu and Ni, Junfeng and Zhu, Song-Chun and Huang, Siyuan},
  journal={arXiv preprint arXiv:2509.17647},
  year={2025},
  eprint={2509.17647},
  archivePrefix={arXiv},
  url={https://arxiv.org/abs/2509.17647}
}

@inproceedings{jiang2023chairs,
  title={Full-Body Articulated Human-Object Interaction},
  author={Jiang, Nan and Liu, Tengyu and Cao, Zhexuan and Cui, Jieming and Zhang, Zhiyuan and Chen, Yixin and Wang, He and Zhu, Yixin and Huang, Siyuan},
  booktitle={Proceedings of the IEEE/CVF International Conference on Computer Vision},
  year={2023}
}

@inproceedings{xue2026rhino,
  title={{RHINO}: Reconstructing Human Interactions with Novel Objects from Monocular Videos},
  author={Xue, Lixin and Zheng, Chengwei and Paschalidis, Georgios and Guo, Chen and Kaufmann, Manuel and Zarate, Juan and Tzionas, Dimitrios},
  booktitle={Proceedings of the IEEE/CVF Conference on Computer Vision and Pattern Recognition},
  year={2026},
  eprint={2605.17014},
  archivePrefix={arXiv},
  url={https://arxiv.org/abs/2605.17014}
}

@inproceedings{taheri2022goal,
  title={{GOAL}: Generating {4D} Whole-Body Motion for Hand-Object Grasping},
  author={Taheri, Omid and Choutas, Vasileios and Black, Michael J. and Tzionas, Dimitrios},
  booktitle={Proceedings of the IEEE/CVF Conference on Computer Vision and Pattern Recognition},
  year={2022}
}

@inproceedings{zhang2022couch,
  title={{COUCH}: Towards Controllable Human-Chair Interactions},
  author={Zhang, Xiaohan and Bhatnagar, Bharat Lal and Guzov, Vladimir and Starke, Sebastian and Pons-Moll, Gerard},
  booktitle={European Conference on Computer Vision},
  year={2022}
}

@inproceedings{xu2023interdiff,
  title={{InterDiff}: Generating {3D} Human-Object Interactions with Physics-Informed Diffusion},
  author={Xu, Sirui and Li, Zhengyuan and Wang, Yu-Xiong and Gui, Liang-Yan},
  booktitle={Proceedings of the IEEE/CVF International Conference on Computer Vision},
  year={2023}
}

@inproceedings{li2024chois,
  title={Controllable Human-Object Interaction Synthesis},
  author={Li, Jiaman and Clegg, Alexander and Mottaghi, Roozbeh and Wu, Jiajun and Puig, Xavier and Liu, C. Karen},
  booktitle={European Conference on Computer Vision},
  year={2024}
}

@inproceedings{hassan2023synthesizing,
  title={Synthesizing Physical Character-Scene Interactions},
  author={Hassan, Mohamed and Guo, Yunrong and Wang, Tingwu and Black, Michael and Fidler, Sanja and Peng, Xue Bin},
  booktitle={ACM SIGGRAPH 2023 Conference Proceedings},
  year={2023}
}

@inproceedings{zhang2026giraf,
  title={{GIRAF}: Towards Generalizable Human Interactions with Articulated Objects},
  author={Zhang, Xiaohan and Starke, Sebastian and Winkler, Alexander and Bogo, Federica and Aroudj, Samir and Ye, Yuting},
  booktitle={Proceedings of the IEEE/CVF Conference on Computer Vision and Pattern Recognition Workshops},
  year={2026},
  eprint={2607.07880},
  archivePrefix={arXiv},
  url={https://arxiv.org/abs/2607.07880}
}

@article{peng2018deepmimic,
  title={DeepMimic: Example-Guided Deep Reinforcement Learning of Physics-Based Character Skills},
  author={Peng, Xue Bin and Abbeel, Pieter and Levine, Sergey and van de Panne, Michiel},
  journal={ACM Transactions on Graphics},
  volume={37},
  number={4},
  year={2018},
  eprint={1804.02717},
  archivePrefix={arXiv},
  url={https://arxiv.org/abs/1804.02717}
}

@article{peng2021amp,
  title={{AMP}: Adversarial Motion Priors for Stylized Physics-Based Character Control},
  author={Peng, Xue Bin and Ma, Ze and Abbeel, Pieter and Levine, Sergey and Kanazawa, Angjoo},
  journal={ACM Transactions on Graphics},
  volume={40},
  number={4},
  year={2021},
  eprint={2104.02180},
  archivePrefix={arXiv},
  url={https://arxiv.org/abs/2104.02180}
}

@inproceedings{yuan2021simpoe,
  title={{SimPoE}: Simulated Character Control for {3D} Human Pose Estimation},
  author={Yuan, Ye and Wei, Shih-En and Simon, Tomas and Kitani, Kris and Saragih, Jason},
  booktitle={Proceedings of the IEEE/CVF Conference on Computer Vision and Pattern Recognition},
  year={2021}
}

@inproceedings{luo2023perpetual,
  title={Perpetual Humanoid Control for Real-time Simulated Avatars},
  author={Luo, Zhengyi and Cao, Jinkun and Winkler, Alexander and Kitani, Kris and Xu, Weipeng},
  booktitle={Proceedings of the IEEE/CVF International Conference on Computer Vision},
  year={2023}
}

@inproceedings{luo2024universal,
  title={Universal Humanoid Motion Representations for Physics-Based Control},
  author={Luo, Zhengyi and Cao, Jinkun and Merel, Josh and Winkler, Alexander and Huang, Jing and Kitani, Kris and Xu, Weipeng},
  booktitle={International Conference on Learning Representations},
  year={2024},
  eprint={2310.04582},
  archivePrefix={arXiv},
  url={https://arxiv.org/abs/2310.04582}
}

@article{tessler2024maskedmimic,
  title={{MaskedMimic}: Unified Physics-Based Character Control Through Masked Motion Inpainting},
  author={Tessler, Chen and Guo, Yunrong and Nabati, Ofir and Chechik, Gal and Peng, Xue Bin},
  journal={ACM Transactions on Graphics},
  volume={43},
  number={6},
  year={2024}
}

@inproceedings{wang2025skillmimic,
  title={{SkillMimic}: Learning Basketball Interaction Skills from Demonstrations},
  author={Wang, Yinhuai and Zhao, Qihan and Yu, Runyi and Tsui, Hok Wai and Zeng, Ailing and Lin, Jing and Luo, Zhengyi and Yu, Jiwen and Li, Xiu and Chen, Qifeng and Zhang, Jian and Zhang, Lei and Tan, Ping},
  booktitle={Proceedings of the IEEE/CVF Conference on Computer Vision and Pattern Recognition},
  year={2025}
}

@inproceedings{luo2024omnigrasp,
  title={Omnigrasp: Grasping Diverse Objects with Simulated Humanoids},
  author={Luo, Zhengyi and Cao, Jinkun and Christen, Sammy and Winkler, Alexander and Kitani, Kris and Xu, Weipeng},
  booktitle={Advances in Neural Information Processing Systems},
  volume={37},
  pages={2161--2184},
  year={2024},
  doi={10.52202/079017-0070}
}

@inproceedings{zhang2024artigrasp,
  title={{ArtiGrasp}: Physically Plausible Synthesis of Bi-Manual Dexterous Grasping and Articulation},
  author={Zhang, Hui and Christen, Sammy and Fan, Zicong and Zheng, Luocheng and Hwangbo, Jemin and Song, Jie and Hilliges, Otmar},
  booktitle={International Conference on 3D Vision (3DV)},
  year={2024}
}

@article{zhang2026dragmesh2,
  title={{DragMesh-2}: Physically Plausible Dexterous Hand-Object Interaction with Articulated Objects},
  author={Zhang, Tianshan and Duan, Yijia and Li, Yanjun and Zhang, Zeyu and Tang, Hao},
  journal={arXiv preprint arXiv:2606.15133},
  year={2026},
  eprint={2606.15133},
  archivePrefix={arXiv},
  url={https://arxiv.org/abs/2606.15133}
}

@inproceedings{mandi2026dexmachina,
  title={{DexMachina}: Functional Retargeting for Bimanual Dexterous Manipulation},
  author={Mandi, Zhao and Hou, Yifan and Fox, Dieter and Narang, Yashraj and Mandlekar, Ajay and Song, Shuran},
  booktitle={International Conference on Machine Learning (ICML)},
  year={2026},
  eprint={2505.24853},
  archivePrefix={arXiv},
  url={https://arxiv.org/abs/2505.24853}
}

@inproceedings{kerr2024rsrd,
  title={Robot See Robot Do: Imitating Articulated Object Manipulation with Monocular {4D} Reconstruction},
  author={Kerr, Justin and Kim, Chung Min and Wu, Mingxuan and Yi, Brent and Wang, Qianqian and Goldberg, Ken and Kanazawa, Angjoo},
  booktitle={Conference on Robot Learning (CoRL)},
  year={2024},
  eprint={2409.18121},
  archivePrefix={arXiv},
  url={https://arxiv.org/abs/2409.18121}
}

@inproceedings{deng2026vlmrmd,
  title={Human-Object Interaction via Automatically Designed {VLM}-Guided Motion Policy},
  author={Deng, Zekai and Shi, Ye and Ji, Kaiyang and Xu, Lan and Huang, Shaoli and Wang, Jingya},
  booktitle={International Conference on Learning Representations},
  year={2026},
  eprint={2503.18349},
  archivePrefix={arXiv},
  url={https://arxiv.org/abs/2503.18349}
}

@article{zhao2025resmimic,
  title={{ResMimic}: From General Motion Tracking to Humanoid Whole-body Loco-Manipulation via Residual Learning},
  author={Zhao, Siheng and Ze, Yanjie and Wang, Yue and Liu, C. Karen and Abbeel, Pieter and Shi, Guanya and Duan, Rocky},
  journal={arXiv preprint arXiv:2510.05070},
  year={2025},
  eprint={2510.05070},
  archivePrefix={arXiv},
  url={https://arxiv.org/abs/2510.05070}
}

@article{zhu2026chord,
  title={Learning Dexterous Manipulation Using Contact Wrench Guidance From Human Demonstration},
  author={Zhu, Xinghao and Liu, Zixi and Jain, Shalin and Li, Chenran and Noori, Milad and Lin, Michael Andres and Zhao, Huihua and Welsh, John and Verghese, Mrinal and Liu, Wei and Wang, Tingwu and Da, Xingye and Luo, Zhengyi and Kulkarni, Vishal and Bhatti, Naema and Zhu, Yuke and Fan, Linxi and Wen, Bowen and Xu, Danfei and Pouya, Soha and Chang, Yan},
  journal={arXiv preprint arXiv:2607.00033},
  year={2026},
  eprint={2607.00033},
  archivePrefix={arXiv},
  url={https://arxiv.org/abs/2607.00033}
}

@inproceedings{yu2026wristmimic,
  title={{WristMimic}: Full-Body Humanoid Control with Wrist-Guided Manipulation},
  author={Yu, Wongyun and Kim, Youngwoon and Cho, Minsu},
  booktitle={European Conference on Computer Vision},
  year={2026}
}

@article{wu2026sugar,
  title={SUGAR: A Scalable Human-Video-Driven Generalizable Humanoid Loco-Manipulation Learning Framework},
  author={Wu, Tianshu and Kong, Xiangqi and Chen, Yue and Yu, Qize and Ye, Hang and Li, Jia and Wang, Yizhou and Dong, Hao},
  journal={arXiv preprint arXiv:2605.20373},
  year={2026},
  eprint={2605.20373},
  archivePrefix={arXiv},
  url={https://arxiv.org/abs/2605.20373}
}

@inproceedings{karaev2025cotracker3,
  title={{CoTracker3}: Simpler and Better Point Tracking by Pseudo-Labelling Real Videos},
  author={Karaev, Nikita and Makarov, Yuri and Wang, Jianyuan and Neverova, Natalia and Vedaldi, Andrea and Rupprecht, Christian},
  booktitle={Proceedings of the IEEE/CVF International Conference on Computer Vision},
  pages={6013--6022},
  year={2025},
  url={https://openaccess.thecvf.com/content/ICCV2025/html/Karaev_CoTracker3_Simpler_and_Better_Point_Tracking_by_Pseudo-Labelling_Real_Videos_ICCV_2025_paper.html}
}

@article{lin2025depthanything3,
  title={Depth Anything 3: Recovering the Visual Space from Any Views},
  author={Lin, Haotong and Chen, Sili and Liew, Jun Hao and Chen, Donny Y. and Li, Zhenyu and Shi, Guang and Feng, Jiashi and Kang, Bingyi},
  journal={arXiv preprint arXiv:2511.10647},
  year={2025},
  eprint={2511.10647},
  archivePrefix={arXiv},
  primaryClass={cs.CV},
  url={https://arxiv.org/abs/2511.10647}
}

@inproceedings{ravi2024sam2,
  title={{SAM} 2: Segment Anything in Images and Videos},
  author={Ravi, Nikhila and Gabeur, Valentin and Hu, Yuan-Ting and Hu, Ronghang and Ryali, Chaitanya and Ma, Tengyu and Khedr, Haitham and R{\"a}dle, Roman and Rolland, Chlo{\'e} and Gustafson, Laura and Mintun, Eric and Pan, Junting and Alwala, Kalyan Vasudev and Carion, Nicolas and Wu, Chao-Yuan and Girshick, Ross and Doll{\'a}r, Piotr and Feichtenhofer, Christoph},
  booktitle={International Conference on Learning Representations},
  year={2025},
  eprint={2408.00714},
  archivePrefix={arXiv},
  primaryClass={cs.CV},
  url={https://proceedings.iclr.cc/paper_files/paper/2025/hash/45c1f6a8cbf2da59ebf2c802b4f742cd-Abstract-Conference.html}
}

@article{carion2025sam3,
  title={{SAM} 3: Segment Anything with Concepts},
  author={Carion, Nicolas and Gustafson, Laura and Hu, Yuan-Ting and Debnath, Shoubhik and Hu, Ronghang and Suris, D{\'i}dac and Ryali, Chaitanya and Alwala, Kalyan Vasudev and Khedr, Haitham and Huang, Andrew and Lei, Jie and Ma, Tengyu and Guo, Baishan and Kalla, Arpit and Marks, Markus and Greer, Joseph and Wang, Meng and Sun, Peize and R{\"a}dle, Roman and Afouras, Triantafyllos and Mavroudi, Effrosyni and Xu, Katherine and Wu, Tsung-Han and Zhou, Yu and Momeni, Liliane and Hazra, Rishi and Ding, Shuangrui and Vaze, Sagar and Porcher, Francois and Li, Feng and Li, Siyuan and Kamath, Aishwarya and Cheng, Ho Kei and Doll{\'a}r, Piotr and Ravi, Nikhila and Saenko, Kate and Zhang, Pengchuan and Feichtenhofer, Christoph},
  journal={arXiv preprint arXiv:2511.16719},
  year={2025},
  eprint={2511.16719},
  archivePrefix={arXiv},
  primaryClass={cs.CV},
  url={https://arxiv.org/abs/2511.16719}
}

@article{yang2026sam3dbody,
  title={{SAM} {3D} Body: Robust Full-Body Human Mesh Recovery},
  author={Yang, Xitong and Kukreja, Devansh and Pinkus, Don and Sagar, Anushka and Fan, Taosha and Park, Jinhyung and Shin, Soyong and Cao, Jinkun and Liu, Jiawei and Ugrinovic, Nicolas and Feiszli, Matt and Malik, Jitendra and Doll{\'a}r, Piotr and Kitani, Kris},
  journal={arXiv preprint arXiv:2602.15989},
  year={2026},
  eprint={2602.15989},
  archivePrefix={arXiv},
  primaryClass={cs.CV},
  url={https://arxiv.org/abs/2602.15989}
}

@article{schulman2017ppo,
  title={Proximal Policy Optimization Algorithms},
  author={Schulman, John and Wolski, Filip and Dhariwal, Prafulla and Radford, Alec and Klimov, Oleg},
  journal={arXiv preprint arXiv:1707.06347},
  year={2017},
  eprint={1707.06347},
  archivePrefix={arXiv},
  primaryClass={cs.LG},
  url={https://arxiv.org/abs/1707.06347}
}

@inproceedings{nasiriany2024robocasa,
  title={{RoboCasa}: Large-Scale Simulation of Household Tasks for Generalist Robots},
  author={Nasiriany, Soroush and Maddukuri, Abhiram and Zhang, Lance and Parikh, Adeet and Lo, Aaron and Joshi, Abhishek and Mandlekar, Ajay and Zhu, Yuke},
  booktitle={Robotics: Science and Systems},
  year={2024},
  doi={10.15607/RSS.2024.XX.050}
}

@misc{minimax2026h3,
  author       = {{MiniMax}},
  title        = {{MiniMax H3}},
  howpublished = {\url{https://github.com/MiniMax-AI/MiniMax-H3}},
  year         = {2026}
}

@misc{qwen2026qwen37plus,
  author       = {{Qwen Team}},
  title        = {{Qwen3.7-Plus}},
  howpublished = {Alibaba Cloud Model Studio},
  url          = {https://help.aliyun.com/en/model-studio/qwen3-7-plus},
  year         = {2026}
}

@misc{qwen2026qwen38flashnext,
  author       = {{Qwen Team}},
  title        = {{Qwen3.8-Flash-Next}},
  howpublished = {Official model card},
  url          = {https://huggingface.co/Qwen/Qwen3.8-Flash-Next},
  year         = {2026}
}

@misc{innerscenepeople,
  title={Free {3D} People and Scale Figures},
  author={{Innerscene}},
  howpublished={Innerscene 3D Parts Library},
  year={n.d.},
  url={https://www.innerscene.com/tools/library/3d-parts/tag/people}
}

@inproceedings{wu2024hoifhli,
  title={Human-Object Interaction from Human-Level Instructions},
  author={Wu, Zhen and Li, Jiaman and Xu, Pei and Liu, C. Karen},
  booktitle={Proceedings of the IEEE/CVF International Conference on Computer Vision},
  pages={11176--11186},
  year={2025},
  doi={10.1109/ICCV51701.2025.01040}
}

@article{viterbi1967error,
  title   = {Error Bounds for Convolutional Codes and an Asymptotically Optimum Decoding Algorithm},
  author  = {Viterbi, Andrew J.},
  journal = {IEEE Transactions on Information Theory},
  volume  = {13},
  number  = {2},
  pages   = {260--269},
  year    = {1967}
}

@inproceedings{makoviychuk2021isaac,
  title={Isaac Gym: High Performance {GPU}-Based Physics Simulation for Robot Learning},
  author={Makoviychuk, Viktor and Wawrzyniak, Lukasz and Guo, Yunrong and Lu, Michelle and Storey, Kier and Macklin, Miles and Hoeller, David and Rudin, Nikita and Allshire, Arthur and Handa, Ankur and State, Gavriel},
  booktitle={NeurIPS Datasets and Benchmarks Track},
  year={2021},
  url={https://arxiv.org/abs/2108.10470}
}

@inproceedings{pavlakos2019expressive,
  title={Expressive Body Capture: {3D} Hands, Face, and Body from a Single Image},
  author={Pavlakos, Georgios and Choutas, Vasileios and Ghorbani, Nima and Bolkart, Timo and Osman, Ahmed A. A. and Tzionas, Dimitrios and Black, Michael J.},
  booktitle={Proceedings of the IEEE/CVF Conference on Computer Vision and Pattern Recognition},
  pages={10967--10977},
  year={2019},
  doi={10.1109/CVPR.2019.01123}
}

@inproceedings{pavlakos2024hamer,
  title={Reconstructing Hands in {3D} with Transformers},
  author={Pavlakos, Georgios and Shan, Dandan and Radosavovic, Ilija and Kanazawa, Angjoo and Fouhey, David and Malik, Jitendra},
  booktitle={Proceedings of the IEEE/CVF Conference on Computer Vision and Pattern Recognition},
  pages={9826--9836},
  year={2024},
  url={https://openaccess.thecvf.com/content/CVPR2024/html/Pavlakos_Reconstructing_Hands_in_3D_with_Transformers_CVPR_2024_paper.html}
}

@inproceedings{xu2022vitpose,
  title={{ViTPose}: Simple Vision Transformer Baselines for Human Pose Estimation},
  author={Xu, Yufei and Zhang, Jing and Zhang, Qiming and Tao, Dacheng},
  booktitle={Advances in Neural Information Processing Systems},
  volume={35},
  pages={38571--38584},
  year={2022},
  doi={10.52202/068431-2795}
}

@inproceedings{potamias2025wilor,
  title={{WiLoR}: End-to-End {3D} Hand Localization and Reconstruction in-the-Wild},
  author={Potamias, Rolandos Alexandros and Zhang, Jinglei and Deng, Jiankang and Zafeiriou, Stefanos},
  booktitle={Proceedings of the IEEE/CVF Conference on Computer Vision and Pattern Recognition},
  pages={12242--12254},
  year={2025},
  doi={10.1109/CVPR52734.2025.01143}
}

@article{romero2017mano,
  title={Embodied Hands: Modeling and Capturing Hands and Bodies Together},
  author={Romero, Javier and Tzionas, Dimitrios and Black, Michael J.},
  journal={ACM Transactions on Graphics},
  volume={36},
  number={6},
  pages={245:1--245:17},
  year={2017},
  doi={10.1145/3130800.3130883}
}
}

\clearpage
\renewcommand\thesection{\Alph{section}}
\renewcommand\thefigure{\Alph{section}\arabic{figure}}
\renewcommand\thetable{\Alph{section}\arabic{table}}
\renewcommand\theequation{\Alph{section}\arabic{equation}}
\renewcommand\theHsection{\Alph{section}}
\renewcommand\theHfigure{\Alph{section}\arabic{figure}}
\renewcommand\theHtable{\Alph{section}\arabic{table}}
\renewcommand\theHequation{\Alph{section}\arabic{equation}}
\setcounter{section}{0}
\setcounter{figure}{0}
\setcounter{table}{0}
\setcounter{equation}{0}
\counterwithin*{figure}{section}
\counterwithin*{table}{section}
\counterwithin*{equation}{section}

\twocolumn[
  \centering
  \Large
  \textbf{\thetitle}\\
  \vspace{0.5em}Appendix\\
  \vspace{1.0em}
]

\vspace{2ex}
\noindent
This appendix provides further implementation and evaluation details for \om{}.
\cref{app:implementation} details the reconstruction pipeline and
physics refiner, including the simulation setup, policy observations,
rewards, and training.
\cref{app:experimental-details} introduces the benchmarks and
RoboCasa video generation, then specifies the baseline adaptations,
evaluation protocol, and metrics.
\cref{app:additional-results} presents further reconstruction and
physics comparisons, covers computational cost and object-goal policy
distillation, shows a qualitative Unitree G1 demonstration in
simulation, and analyzes failed rollouts.
Finally, \cref{app:limitations} outlines the limitations of our
approach, directions for future work, and broader impact.

\section{Implementation Details}
\label{app:implementation}

\subsection{Contact-Aware Articulated Reconstruction}
\label{app:reconstruction}

Given a video-asset pair $(V,\mathcal{A})$, reconstruction estimates the reference $\widehat{\mathcal{R}}_{1:T}$, where $\mathcal{A}$ defines the canonical geometry and kinematics.
The object state consists of a world 6D pose $\widehat{T}^{o}_t$, the revolute and prismatic joint values $\widehat{q}_t$, and a scalar scale $\widehat{s}$.
Our reconstruction pipeline initializes the scene and object, tracks the moving joint over the video, and aligns the hand with the object's moving contact surface.

\emph{Stage 1: Scene and object initialization (domain-dependent).}
From the video, the asset, and the preprocessed depth, human pose, and mask cues, this stage establishes the world frame, camera parameters, and initial object root pose $T^o_t$ and scale $s$.
On generated RoboCasa~\citep{nasiriany2024robocasa} videos, the camera, floor height, object root pose, and scale are known from the case file and stay fixed.
On real D3D-HOI~\citep{xu2021d3dhoi} videos, neither the camera nor the object pose is given, so this stage estimates a gravity-aligned floor plane and optimizes the initial object translation, scale, and yaw.

\emph{Stage 2: Trajectory-level articulation tracking.}
This stage takes the 3D lifted point tracks $\{X_{t,i}\}$ on the moving part, the part silhouettes, the asset kinematics, and the VLM contact onset $t_0$.
With $T^o_t$ and $s$ fixed, it initializes the joint trajectory from the silhouettes and fits $q_{1:T}$ to the lifted tracks under joint limits and temporal regularization.
We then refine the tracks and joint trajectory, and smooth the recovered joint states before export.

\emph{Stage 3: Contact-aware HOI alignment.}
We hold the recovered articulation $q_{1:T}$ fixed throughout alignment.
On D3D-HOI, we optimize a sequence-wide correction to the object root pose and scale for 200 iterations with the initial SMPL-X motion fixed.
The contact, penetration, and grasp terms use the VLM contact cues $(\mathcal{C}, \mathcal{F}_t)$ and canonical contact surface $\mathcal{S}_{\ell}$. The objective also includes a support-plane term.
On RoboCasa, the object root pose and scale stay at their scene layout values.

On both datasets, we hold the object state, human root trajectory, body shape, and remaining body joints fixed and refine the manipulating arm and hand for 800 iterations.
We parameterize the clavicle, shoulder, elbow, wrist, and finger rotations as residual cubic B-splines, $\theta_t=\bar{\theta}_t+\sum_k B_k(t)c_k$, where $\bar{\theta}_t$ is the initial pose, $B_k$ are uniform cubic bases, and $c_k$ are the optimized control points.
The knots are spaced 16 frames apart for the arm rotations and 12 frames apart for the fingers.
The canonical contact region $\mathcal{S}_{\ell}$ is attached to the active link and follows it as $\mathcal{S}_{\ell,t}=\mathcal{T}_{\mathrm{cam}}(T^o_t,s,q_t,\ell)\,\mathcal{S}_{\ell}$.
The contact loss pulls the VLM-identified finger regions toward this surface, and the penetration loss penalizes the deepest penetrating hand vertices.
The grasp energy of HOI-FHLI~\citep{wu2024hoifhli} rewards finger-surface proximity and force closure while penalizing penetration and implausible finger poses.
These three terms act only on the manipulating hand during annotated contact frames, whereas the velocity and acceleration regularizer covers the full sequence.

\paragraph{Preprocessing and dataset-specific details.}
Real and generated videos share the same observation extractors but differ in how the camera, object, and active joint are obtained. Details are listed as follows.

\emph{Shared observations.}
We estimate camera-frame human meshes with SAM 3D Body~\citep{yang2026sam3dbody}, convert them to SMPL-X~\citep{pavlakos2019expressive}, and smooth pose and shape with five passes of temporal blending.
Depth Anything 3~\citep{lin2025depthanything3} estimates per-frame depth in streaming chunks of eight frames with four-frame overlap.
As monocular depth has no reliable metric scale, we fit a single robust offset to the depth rendered from the SMPL-X body over the human-mask pixels of all frames and apply it to the whole video, which preserves relative depth while placing the sequence in the metric frame of the body.
CoTracker3~\citep{karaev2025cotracker3} tracks points on the moving part: we query an $11\times11$ grid inside the eroded part mask every ten frames to catch points entering view later, and keep the visibility flags.
We lift each visible track to the camera frame as $X_{t,i}=D_t(u_{t,i})K_t^{-1}\bar{u}_{t,i}$, where $\bar{u}_{t,i}$ is the homogeneous pixel coordinate.
SAM 3~\citep{carion2025sam3} provides the human masks for depth alignment and the floor mask.

\emph{Real videos (D3D-HOI).}
The benchmark provides the matched CAD and moving joint but neither the camera nor the object pose, so we estimate both.
SAM 2~\citep{ravi2024sam2} propagates the object and moving-part masks from the keyframe annotations.
The gravity-aligned world frame comes from the plane fitted to the SAM 3 floor depth in the first frame, shifted along its normal until the SMPL-X feet rest on it in frames where both feet move slowly (below $0.25$\,m/s for at least eight frames).
We initialize the object at the frame with the largest moving-part mask among those, sampled every five frames, after the tracked points start to move.
The scale follows from matching the CAD extent to the depth inside the mask.
We search 36 gravity-aligned yaw hypotheses from $-90^\circ$ to $85^\circ$ in $5^\circ$ steps, solve the translation from mask and depth, and refine translation, scale, and yaw for 200 iterations under the mask loss and a human-position prior.
For objects standing on the floor, a soft support term pulls the lowest CAD vertices onto the floor.

\emph{Generated videos (RoboCasa).}
As the scene is known, we estimate neither the camera nor the object pose.
The case file gives the camera parameters, floor height, and the target fixture's root pose, scale, and initial joint value, which stay fixed while only the joint trajectory and the human motion are optimized.
If the scene contains several instances of the same category, Qwen3.7-Plus~\citep{qwen2026qwen37plus} returns a box around the moving part in the first frame. We prompt SAM 2 with this box and select the candidate whose projected mask has the highest IoU with the resulting mask.
The moving-part mask is propagated from the same box, and floor segmentation is skipped because the scene defines the ground.

\begin{figure*}[t]
  \centering
  \includegraphics[width=0.75\textwidth]{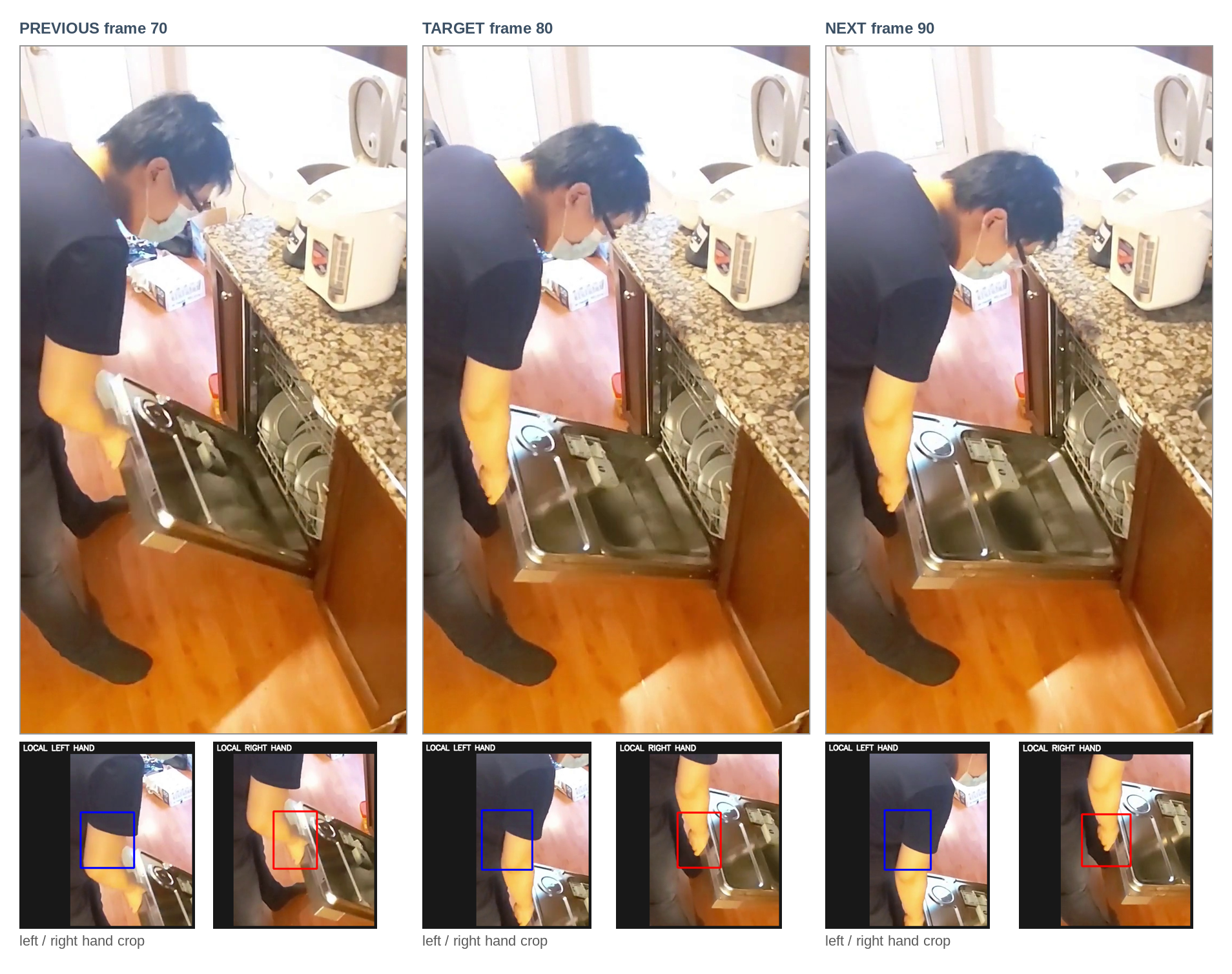}
  \vspace{-1.0em}
  \caption{\textbf{Visual input of one contact query.}
  The query frame (80) is shown with the previous and the next queried frame, and each frame contributes the full-scene image and the left and right hand crops.
  The boxes come from the projected SMPL-X hand joints and mark an estimated hand position only, and the left box here lies on the sleeve.
  The model labels the right thumb, index finger, and middle finger as touching the dishwasher door.}
  \vspace{-0.4em}
  \label{fig:contact-visual-prompt}
\end{figure*}

\paragraph{VLM contact labeling.}
Qwen3.8-Flash-Next~\citep{qwen2026qwen38flashnext} labels which fingers of which hand touch the moving part in two stages using the prompts in \cref{fig:contact-prompt}.
In the first stage, it takes uniformly sampled full-scene frames with timestamps and predicts a coarse interval-level role for each hand (active, passive, none, or uncertain).
In the second stage, we query keyframes at a regular temporal stride (shortened for short clips); each query supplies the target frame and its neighboring queried frames, pairing the full-scene image with projected left- and right-hand crops from the SAM 3D Body SMPL-X joints (\cref{fig:contact-visual-prompt}).
Conditioned on the first-stage assessment, which it overrides when local visual evidence disagrees, the model outputs a per-hand role and per-finger contact booleans, where a finger counts as touching only if its hand is active.
Each label holds until the next query.
The labeled fingers define the participating finger segments $\mathcal{F}_t$, and frames with at least one labeled finger form the contact frames $\mathcal{C}$.
Each labeled hand is assigned to the nearest object part and smoothed over time, and the link's contact region is the vertex set stored with the asset.

\subsection{Articulation-Guided Physics Refinement}
\label{app:physics}

To convert kinematic references into physically executable skills without learning whole-body balance from scratch, \om{} builds a residual policy on top of the frozen pretrained PHC-X whole-body tracker~\citep{luo2024omnigrasp} with articulated fingers.
At each control step, the frozen tracker outputs nominal proportional-derivative (PD) joint targets $u^{\mathrm{ref}}_t$ and the learned residual policy predicts bounded corrections $\delta_t \in [-1,1]^{153}$ across the body and finger joints.
The humanoid tracks the combined targets $u^{\mathrm{pd}}_t = u^{\mathrm{ref}}_t + \sigma\,\delta_t$ using the PD gains of PHC-X, with $\sigma$ scaling the residual.
The simulator initializes the object's joints from the reference. During the rollout, the joints are passive, with no joint drives or actuators, and the humanoid actuates them through contact.
Below we detail the simulation parameters, the 1992-dimensional policy observation, the reward formulation, and the PPO training protocol.

\paragraph{Simulation.}
The SMPL-X~\citep{pavlakos2019expressive} humanoid with articulated fingers and the articulated asset are simulated in Isaac Gym~\citep{makoviychuk2021isaac} at 60\,Hz with two substeps, four position solver iterations, one velocity solver iteration, and a contact offset of 2\,mm. The policy acts at 30\,Hz.
Objects have a density of 1000\,kg/m$^3$ and no restitution, and the object and ground friction coefficients are both 1.0.

\paragraph{Policy.}
At each control step, the policy observes the current humanoid and object state, the reference humanoid DOF, the reference object state and its error to the current object state, the action of the frozen PHC-X tracker~\citep{luo2024omnigrasp}, the fingertip contact forces, and the poses of the finger segments and wrists relative to the object, which form a 1992-D observation (\cref{tab:policy-observations}).
\begin{table}[t]
  \caption{\textbf{Policy observations.} The tracker action is zero for the direct-policy ablation.}
  \vspace{-0.4em}
  \label{tab:policy-observations}

  \centering
  \footnotesize
  \setlength\tabcolsep{4pt}
  \begin{tabular*}{\linewidth}{l@{\extracolsep{\fill}}r}
    \toprule
    Term & Dim. \\
    \midrule
    Humanoid state (bodies, DOF position and velocity) & 1084 \\
    Object state (links, active joint) & 31 \\
    Tracker action & 153 \\
    Reference humanoid DOF & 153 \\
    Reference object state and its tracking error & 61 \\
    Fingertip contact forces & 30 \\
    Finger and wrist poses relative to the object & 480 \\
    \midrule
    Total & 1992 \\
    \bottomrule
  \end{tabular*}
\end{table}

\paragraph{Rewards.}
The body and object tracking rewards combine into one exponential of weighted costs,
\begin{equation}
r^{\mathrm{body}}_t\,r^{\mathrm{obj}}_t=\exp\!\Big(-\sum_{k}\lambda_k E_k\Big),
\end{equation}
with the weights $\lambda_k$ in \cref{tab:reward-weights}.
The body costs $E^{h}_{p}$, $E^{h}_{\theta}$, $E^{h}_{v}$, and $E^{h}_{\omega}$ are the position, rotation, linear-velocity, and angular-velocity errors to the reference, averaged over the bodies, with a lower weight on the hand bodies.
The object costs $E^{\mathrm{r}}_{p}$, $E^{\mathrm{r}}_{\theta}$, and $E^{\mathrm{r}}_{v}$ are the pose and velocity errors of the object root, $E^{\ell}_{p}$ and $E^{\ell}_{\theta}$ the pose errors of the active link, and $E_{q}$ and $E_{\dot q}$ the squared errors of the active joint value and velocity.
Inspired by InterMimic's energy regularization~\citep{xu2025intermimic}, we use $E^{h}_{a}$, $E^{\mathrm{r}}_{a}$, and $E_{f}$ to penalize the squared joint acceleration, the squared object acceleration, and the squared total contact force on the humanoid.
On labeled contact frames, the contact reward averages $\exp(-w_{\mathrm{c}}d_{t,j})$ over the proximal, middle, and distal segments of each labeled finger in $\mathcal F_t$, where $w_{\mathrm{c}}$ sets the distance sensitivity and $d_{t,j}$ measures the distance to the contact region of the moving link (\cref{app:reconstruction}).
The ablation without fine-grained contact applies the same kernel to the distance from each labeled hand to the object.
The progress reward credits joint displacement in the opening or closing direction, weights it by active contact, and penalizes overshoot.
We detect opening and closing phases once from the smoothed reference joint trajectory and measure active contact as the fraction $m_t$ of labeled segments within 2\,cm of the object with a contact force above 0.1\,N.
A phase pays progress only after the prior phase reaches half its reference displacement.
\begin{table}[t]
  \caption{\textbf{Reward weights.}}
  \vspace{-0.4em}
  \label{tab:reward-weights}

  \centering
  \footnotesize
  \setlength{\tabcolsep}{3pt}
  \begin{tabular*}{\linewidth}{l@{\extracolsep{\fill}}l}
    \toprule
    Term & Weight \\
    \midrule
    Body $(\lambda^h_p,\lambda^h_\theta,\lambda^h_v,\lambda^h_\omega)$ & (30, 1.5, 0.25, 0.02) \\
    Hand-body weight & 0.2 \\
    Object root $(\lambda^{\mathrm{r}}_p,\lambda^{\mathrm{r}}_\theta,\lambda^{\mathrm{r}}_v)$ & (5, 0.1, 0.1) \\
    Active link $(\lambda^\ell_p,\lambda^\ell_\theta)$ & (5, 0.1) \\
    Active joint $(\lambda_q,\lambda_{\dot q})$ & (6, 0.5) \\
    Energy $(\lambda^h_a,\lambda^{\mathrm{r}}_a,\lambda_f)$ & ($2\times10^{-5}$, $2\times10^{-5}$, $10^{-10}$) \\
    Contact kernel $w_{\mathrm{c}}$ & 10 \\
    Progress $(w_{\mathrm{prog}},\lambda_{\mathrm{o}})$ & (1.5, 0.5) \\
    \bottomrule
  \end{tabular*}
\end{table}

\paragraph{Training.}
In the main experiments, one policy is trained per CAD asset, with its parallel environments distributed over that asset's reference trajectories.
The single-trajectory setting of \cref{tab:physics-single} trains one policy per trajectory.
An episode starts at the first reference frame, or with probability 0.1 at a random reference state in the first half of the trajectory.
It terminates after 300 steps, when the pelvis drops below 0.3\,m, when the mean key-body or object-surface error exceeds 0.5\,m, or when a labeled hand has no contact for ten consecutive steps.
Evaluation disables these terminations and runs every reference from its first to its last frame.
We train with PPO~\citep{schulman2017ppo} using the hyperparameters in \cref{tab:training-hyper} and evaluate the final checkpoint.
\begin{table}[t]
  \caption{\textbf{PPO hyperparameters}, shared by all benchmarks and ablations.}
  \vspace{-0.4em}
  \label{tab:training-hyper}

  \centering
  \footnotesize
  \setlength{\tabcolsep}{4pt}
  \begin{tabularx}{\linewidth}{l>{\raggedright\arraybackslash}X}
    \toprule
    Hyperparameter & Value \\
    \midrule
    Policy network & separate actor and critic MLPs, 1024-1024-512, ReLU \\
    Residual scale $\sigma$ & 1.0 \\
    Exploration standard deviation & $\exp(-2.9)$ (fixed) \\
    Parallel environments per asset & 4096 \\
    Horizon / episode length & 32 / 300 steps \\
    Minibatch size / mini-epochs & 16384 / 6 \\
    Discount $\gamma$ / GAE $\tau$ & 0.99 / 0.95 \\
    Learning rate & $2\times10^{-5}$ (constant) \\
    PPO clip coefficient & 0.2 \\
    Policy updates & 1000 \\
    \bottomrule
  \end{tabularx}
\end{table}

\section{Experimental Details}
\label{app:experimental-details}

\subsection{Benchmarks}

\cref{tab:dataset-cases} summarizes the per-CAD composition of the main physics benchmarks and the object-category breakdown of the per-trajectory evaluation sets.
\begin{table*}[t]
  \caption{\textbf{Composition of the main per-CAD physics benchmarks and per-trajectory evaluation sets.}
  Counts report evaluated trajectories per object category.
  $^*$In the D3D-HOI per-trajectory set, storage furniture comprises 5 prismatic and 5 revolute trajectories.}
  \vspace{-0.4em}
  \label{tab:dataset-cases}

  \centering
  \footnotesize
  \setlength{\tabcolsep}{5pt}
  \renewcommand{\arraystretch}{1.08}
  \begin{subtable}[t]{0.26\textwidth}
    \centering
    \caption{RoboCasa (16 assets)}
    \begin{tabular}{lr}
      \toprule
      \textbf{Category} & \textbf{Per-CAD} \\
      \midrule
      Cabinet door & 19 \\
      Drawer & 14 \\
      Refrigerator & 17 \\
      Microwave & 25 \\
      Oven & 31 \\
      Dishwasher & 22 \\
      \midrule
      \textbf{Total} & \textbf{128} \\
      \bottomrule
    \end{tabular}
  \end{subtable}%
  \hfill
  \begin{subtable}[t]{0.35\textwidth}
    \centering
    \caption{D3D-HOI (16 assets)}
    \begin{tabular}{lrr}
      \toprule
      \textbf{Category} & \textbf{Per-CAD} & \textbf{Per-Traj.} \\
      \midrule
      Dishwasher & 11 & 5 \\
      Laptop & 41 & 5 \\
      Microwave & 36 & 5 \\
      Oven & 11 & 5 \\
      Refrigerator & 80 & 5 \\
      Storage furniture$^*$ & 32 & 10 \\
      Trash can & 16 & 5 \\
      Washing machine & 24 & 5 \\
      \midrule
      \textbf{Total} & \textbf{251} & \textbf{45} \\
      \bottomrule
    \end{tabular}
  \end{subtable}%
  \hfill
  \begin{subtable}[t]{0.35\textwidth}
    \centering
    \caption{ARCTIC (8 assets)}
    \begin{tabular}{lrr}
      \toprule
      \textbf{Category} & \textbf{Per-CAD} & \textbf{Per-Traj.} \\
      \midrule
      Box & 20 & 4 \\
      Capsule machine & 22 & 4 \\
      Espresso machine & 22 & 4 \\
      Laptop & 24 & 4 \\
      Microwave & 21 & 4 \\
      Mixer & 23 & 4 \\
      Notebook & 22 & 4 \\
      Waffle iron & 24 & 4 \\
      Ketchup & -- & 4 \\
      Phone & -- & 4 \\
      \midrule
      \textbf{Total} & \textbf{178} & \textbf{40} \\
      \bottomrule
    \end{tabular}
  \end{subtable}
\end{table*}

D3D-HOI~\citep{xu2021d3dhoi} pairs each real RGB sequence with a matched PartNet-Mobility~\citep{xiang2020sapien} CAD model and supports both reconstruction and video-to-physics evaluation.
We exclude three of its 256 videos that show a foot-operated trash-can, as they contain no hand-object interaction, which leaves 253 videos for reconstruction.
The revolute and prismatic joint metrics cover 236 and 17 of these videos, respectively.
For the main per-CAD physics evaluation, excluding two additional clips with negligible articulation yields 251 D3D-HOI trajectories across 16 CAD assets.
The D3D-HOI per-trajectory experiment evaluates 45 cases (\cref{tab:physics-single}).
For RoboCasa~\citep{nasiriany2024robocasa}, we retain 203 of 276 generated videos after filtering and evaluate 128 trajectories across 16 CAD assets in the main per-CAD physics benchmark.
ARCTIC~\citep{fan2023arctic} provides motion-captured references and is used for physics evaluation only.
ARCTIC's main per-CAD physics evaluation uses 178 trajectories across 8 object assets.
The ARCTIC per-trajectory experiment trains one policy per trajectory across 40 captured sequences (\cref{tab:physics-captured}).

\subsection{Asset-Conditioned Generated Videos}
\label{app:generated-video-protocol}

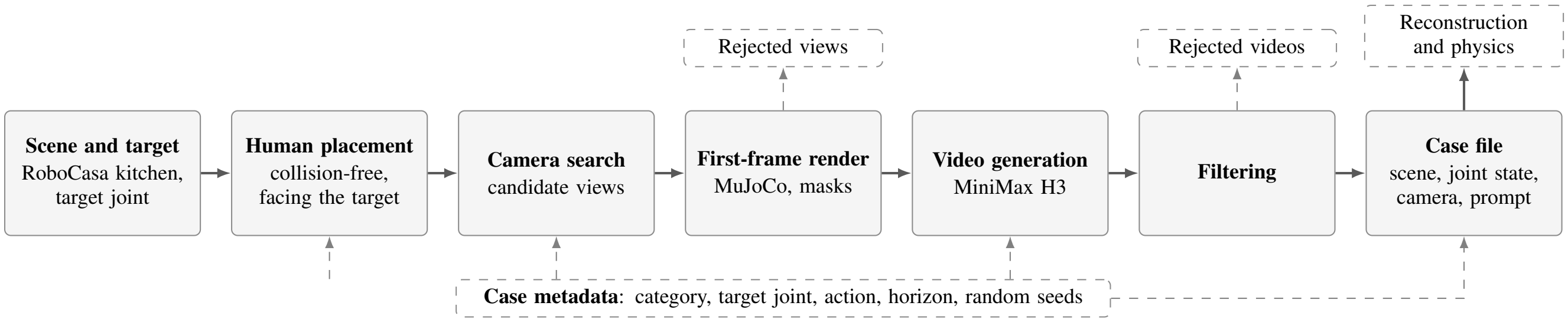
\begin{figure*}[t]
  \centering
  \resizebox{\linewidth}{!}{%
  \begin{tikzpicture}[
    stage/.style={draw=black!45, fill=black!4, rounded corners=2pt,
      align=center, text width=2.1cm, minimum height=1.45cm,
      font=\scriptsize, inner sep=2.5pt},
    aux/.style={draw=black!45, dashed, rounded corners=2pt, align=center,
      font=\scriptsize, inner sep=3pt},
    flow/.style={-{Latex[length=1.6mm]}, thick, black!65},
    side/.style={-{Latex[length=1.4mm]}, black!50, dashed},
    node distance=3.5mm]
    \node[stage] (scene) {\textbf{Scene and target}\\[1pt]
      RoboCasa kitchen,\\target joint};
    \node[stage, right=of scene] (human) {\textbf{Human placement}\\[1pt]
      collision-free,\\facing the target};
    \node[stage, right=of human] (camera) {\textbf{Camera search}\\[1pt]
      candidate views};
    \node[stage, right=of camera] (render) {\textbf{First-frame render}\\[1pt]
      MuJoCo, masks};
    \node[stage, right=of render] (video) {\textbf{Video generation}\\[1pt]
      MiniMax H3};
    \node[stage, right=of video] (screen) {\textbf{Filtering}};
    \node[stage, right=of screen] (manifest) {\textbf{Case file}\\[1pt]
      scene, joint state,\\camera, prompt};
    \draw[flow] (scene) -- (human);
    \draw[flow] (human) -- (camera);
    \draw[flow] (camera) -- (render);
    \draw[flow] (render) -- (video);
    \draw[flow] (video) -- (screen);
    \draw[flow] (screen) -- (manifest);
    \node[aux, below=5mm of $(camera.south)!0.5!(video.south)$, text width=7.4cm] (meta)
      {\textbf{Case metadata}: category, target joint, action, horizon, random seeds};
    \draw[side] (meta.north -| human.south) -- (human.south);
    \draw[side] (meta.north -| camera.south) -- (camera.south);
    \draw[side] (meta.north -| video.south) -- (video.south);
    \draw[side] (meta.east) -| (manifest.south);
    \node[aux, above=5mm of render, text width=2.1cm] (reject1)
      {Rejected views};
    \draw[side] (render.north) -- (reject1.south);
    \node[aux, above=5mm of screen, text width=2.1cm] (reject2)
      {Rejected videos};
    \draw[side] (screen.north) -- (reject2.south);
    \node[aux, above=5mm of manifest, text width=2.1cm] (recon)
      {Reconstruction and physics};
    \draw[flow] (manifest.north) -- (recon.south);
  \end{tikzpicture}}
  \caption{\textbf{Generated-video pipeline on RoboCasa kitchens.}
  Solid arrows denote the per-case data flow, and dashed arrows denote case metadata reused by later stages and the views and videos rejected by filtering.
  Every stage reads the output of the previous one together with the case metadata, so the pipeline scales with the number of scenes and targets.}
  \vspace{-0.4em}
  \label{fig:videogen-workflow}
\end{figure*}

Generated videos provide interaction priors at scale without motion capture.
We build them from the kitchen scenes of RoboCasa~\citep{nasiriany2024robocasa}, whose fixtures are articulated MuJoCo~\citep{todorov2012mujoco} models with known kinematics.
The same asset therefore conditions video generation and defines the reconstruction and simulation model.
As shown in \cref{fig:videogen-workflow}, the pipeline places a person in front of the target fixture, renders the first frame, generates a video from it, and stores the video together with the scene, the target joint, its initial state, and its limits.

\paragraph{Scenes and targets.}
We use 50 kitchen scene cases covering eight articulated targets: cabinet doors, drawers, refrigerator doors, microwave doors, oven doors, oven racks, dishwasher doors, and dishwasher racks.
The scenes come from 12 RoboCasa floor-plan layouts with two style variants each, and a style change alone does not count as a new layout.
Each case specifies the target fixture and joint and is exported with its initial state, so every target has a known pose, joint type, and joint limits.
We generate 276 videos from these scene and target configurations, varying the human identity, placement, camera, and interaction direction while retaining the same asset-conditioned scene metadata.
Most cases start with the person standing in front of the target, and the interaction prompt selects the approach motion from the initial distance to the target: the person remains within arm's reach for nearby placements, takes a small step for intermediate placements, and walks several natural steps for more distant placements.

\paragraph{Human placement.}
We assign one of ten static adult InnerScene assets~\citep{innerscenepeople} to each generated video and sample candidate root positions on a 20\,cm grid around the anchor of the moving part, rank them by their deviation from a horizon-dependent target distance, and discard every candidate whose body bounding box collides with the kitchen.
The selected person faces the target with small offset and yaw perturbations drawn from fixed per-case seeds.
The initial pose is selected from the standing assets, with the prompt specifying a natural reach or approach consistent with the target height and the initial distance.

\paragraph{Camera search and first-frame rendering.}
We place an eye-level camera on a ring around the person and the target, at a side three-quarter view relative to the person's facing direction and at a horizon-dependent radius.
For each case we render twelve candidate views in MuJoCo at $768\times512$ pixels together with the human and target-part masks, and retain the highest-scoring geometrically valid view after a low-resolution pass and a full-resolution check.
A view is scored by whether the full body is visible and sufficiently large, by the visible area and image position of the moving part, and by the relative position and orientation of the person and the target.

\paragraph{Video generation.}
We generate each video with the distilled MiniMax H3 image-to-video model~\citep{minimax2026h3}, conditioned on the selected first frame and a prompt instantiated from the template in \cref{fig:videogen-prompts}.
The generator uses eight inference steps, a fixed seed for each recorded generation request, a spatial resolution of $832\times544$ pixels, 124 frames at 24\,Hz, and a duration of approximately 5.17\,s.
The rendered first frame is resized and encoded to the video resolution before image-to-video inference.

\begin{figure}[t]
  \centering
  \fcolorbox{black!30}{black!4}{\parbox{\dimexpr\linewidth-2\fboxsep-2\fboxrule\relax}{\small
  \textbf{Prompt.}
  A single adult performs a natural full-body kitchen interaction: \textit{\{action\}}.
  Keep the camera fixed, preserve the kitchen layout and the target appliance, show the whole body and the moving articulated part clearly, realistic motion, no cuts.

  \smallskip
  \textbf{Negative prompt.}
  Camera cut, camera motion, extra people, object deformation, changing kitchen, cropped body, hidden target.}}
  \caption{\textbf{Prompt template for generated videos.}
  The italic field is filled with the target action, for example ``pull a drawer open'' or ``open the oven and pull its rack out''.}
  \vspace{-0.4em}
  \label{fig:videogen-prompts}
\end{figure}

\begin{figure*}[t]
  \centering
  \includegraphics[width=\textwidth]{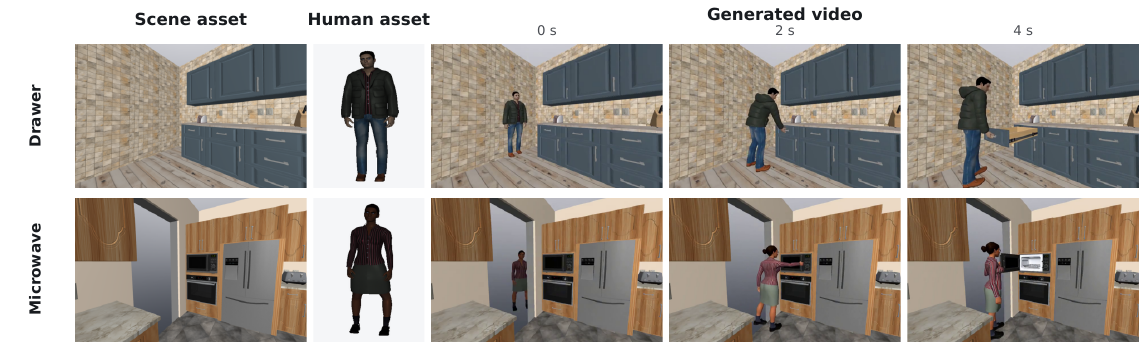}
  \caption{\textbf{Representative generated videos on RoboCasa.}
  Each row shows the source kitchen scene, the original InnerScene avatar, and frames from the generated video at 0, 2, and 4 seconds.}
  \vspace{-0.4em}
  \label{fig:robocasa-gallery}
\end{figure*}

\paragraph{Video Filtering.}
Before reconstruction, Qwen3.8-Flash-Next~\citep{qwen2026qwen38flashnext} filters each generated video against four criteria: the whole body and the target stay in view, the clip has no cut or camera motion, the moving part is the target part, and it moves in the direction of the action only after contact.
Filtering retains 203 of the 276 generated videos (73.6\%) and mainly removes videos with generation artifacts.
\cref{fig:robocasa-gallery} shows the source assets and generated video frames for two RoboCasa cases.

\subsection{Articulated Object Models}

Each articulated asset provides the kinematic tree, the joint types and limits, and the visual and collision geometry.
The visual meshes drive reconstruction and rendering, while the collision geometry, mass, inertia, and joint limits drive simulation.
Both share the same link names, joint axes, and root frame, and joint values are absolute, in radians for revolute joints and meters for prismatic joints.

\subsection{Baseline Implementations}
\label{app:baseline-implementation}

\paragraph{Reconstruction baselines.}
ArtHOI-GS~\citep{huang2026arthoi} and ArtHOI-Hand~\citep{wang2026arthoi} receive the same CAD asset, camera parameters, masks, and initial object pose, scale, and joint state as \om{}, and recover part motion by fitting the CAD joint.
On D3D-HOI, the baselines and \om{} without HOI alignment keep the initial object scale, which the full method can refine.

ArtHOI-GS starts from the same SAM 3D Body estimate, adds HaMeR~\citep{pavlakos2024hamer} hand poses and ViTPose~\citep{xu2022vitpose} keypoints, and infers contact from object motion, keypoints, masks, and depth, which yields no valid keypoints in 51 of the 253 D3D-HOI videos.
It uses CAD meshes on D3D-HOI and converts them to 3D Gaussians for generated videos.
ArtHOI-Hand reconstructs hands with WiLoR~\citep{potamias2025wilor} and MANO~\citep{romero2017mano}, uses the shared per-finger contact labels, and computes its own depth and point tracks, with the depth scale aligned to the shared depth on D3D-HOI.

We also score the released D3D-HOI~\citep{xu2021d3dhoi} object predictions on the 253-video benchmark with the shared evaluator, after mapping them to the shared CAD models and interpolating object motion between annotated frames.

\paragraph{Physics baselines.}
\cref{tab:physics-baseline-scope} summarizes each method's original
task, objects, and embodiment.
We adapt the physics baselines to the same SMPL-X~\citep{pavlakos2019expressive}
humanoid, motion references, articulated assets, and evaluator.
Each method keeps its original learning objective.
We run CoDA's released articulated-object controller, which
extends PhysHOI~\citep{wang2023physhoi} through the shared interface.
\begin{table*}[t]
  \caption{\textbf{Comparison of reconstruction and physics methods in their
  native settings.} (a) We compare reconstruction methods by video input,
  human scope, and whether the object model is recovered or provided. (b) We
  compare physics methods by native task, object type, and embodiment.}
  \vspace{-0.4em}
  \label{tab:physics-baseline-scope}

  \centering
  \small
  \setlength{\tabcolsep}{8pt}

  \textbf{(a) Articulated HOI Reconstruction}\par\vspace{0.25em}
  \begin{adjustbox}{max width=\textwidth}
  \begin{tabular}{llll}
    \toprule
    Method & Video Input & Human Scope & Object Representation \\
    \midrule
    ArtHOI~\citep{wang2026arthoi} & Monocular RGB & Hands (MANO)
      & Recovered articulated mesh \\
    ArtHOI-GS~\citep{huang2026arthoi} & Generated video prior & Full body (SMPL-X)
      & Recovered 3D Gaussians \\
    \midrule
    \textbf{Ours} & Monocular RGB / video prior & Full body (SMPL-X)
      & Provided articulated CAD asset \\
    \bottomrule
  \end{tabular}%
  \end{adjustbox}

  \vspace{0.8em}
  \textbf{(b) Physics Execution}\par\vspace{0.25em}
  {\setlength{\tabcolsep}{8pt}
  \begin{adjustbox}{max width=\textwidth}
  \begin{tabular}{llll}
    \toprule
    Method & Native Task & Object Type & Embodiment \\
    \midrule
    InterMimic~\citep{xu2025intermimic} & Whole-body HOI imitation & Rigid & SMPL-X \\
    HDMI~\citep{weng2025hdmi} & Whole-body control from human videos & Articulated & Unitree G1 \\
    CoDA~\citep{pi2025coda} & Whole-body articulated manipulation & Articulated & SMPL-X \\
    \midrule
    \textbf{Ours} & Video-to-physics execution & Articulated & SMPL-X \\
    \bottomrule
  \end{tabular}%
  \end{adjustbox}
  }
\end{table*}

\subsection{Evaluation Protocol}
\label{app:physics-evaluation-protocol}

Within each interaction source, all methods execute the same references: the \om{} reconstructions of real D3D-HOI~\citep{xu2021d3dhoi} videos and generated RoboCasa videos, or the captured ARCTIC~\citep{fan2023arctic} motions.
All methods are trained with 1000 policy updates, 4096 parallel environments, and a 32-step horizon, and every rollout starts from the first reference frame and is scheduled to run to the last.
A rollout that falls or stops early counts as a failure, and continuous errors are averaged over the valid frames, up to and including the first fall frame when a fall occurs.

\paragraph{Per-CAD policies.}
In the main comparison, each method trains one policy per articulated asset on all of its reference trajectories and is evaluated on these trajectories.
RoboCasa uses 128 trajectories across 16 CAD assets, D3D-HOI uses 251 trajectories across 16 CAD assets, and ARCTIC uses 178 trajectories across 8 object assets.
Each trajectory is evaluated with three seeds, and results are averaged over seeds and then over trajectories with equal weight.

\paragraph{Per-trajectory policies.}
In two per-trajectory experiments, each method trains one policy per trajectory on 45 reconstructed D3D-HOI cases (\cref{tab:physics-single}) and on 40 captured ARCTIC trajectories (\cref{tab:physics-captured}).

\subsection{Metrics}
\label{app:metric-audit}

\paragraph{Articulation progress and success.}
The evaluator segments each reference joint trajectory into movement phases $k$ with start and end frames $(s_k,e_k)$ and joint $j_k$, and computes the phase-duration-weighted mean progress $P$ from the per-phase reference-normalized progress $p_k$:
\begin{equation}
\begin{aligned}
  p_k &=\operatorname{clip}_{[0,1]}\!\left[
  \max_{t\in\mathcal W_k}
  \frac{q^{\mathrm{sim}}_{t,j_k}-q^{\mathrm{ref}}_{s_k,j_k}}
       {q^{\mathrm{ref}}_{e_k,j_k}-q^{\mathrm{ref}}_{s_k,j_k}}
  \right], \\ \quad
  P &=\frac{\sum_k L_kp_k}{\sum_k L_k},\quad L_k=e_k-s_k+1,
\end{aligned}
\end{equation}
where $\mathcal W_k$ is the five-frame window centered on the end of phase $k$.
A rollout succeeds if its phase-duration-weighted mean progress satisfies $P\geq\tau$, it reaches the end of the reference without the pelvis dropping below $0.2$\,m, the mean global MPJPE over 22 body joints is below $0.2$\,m, and the joint error over the reference-active frames is below $1.0$\,rad for revolute joints or $0.10$\,m for prismatic joints.
We use $\tau=0.5$ in the main per-asset evaluation and $\tau=0.8$ in the per-trajectory comparisons (\cref{tab:physics-single,tab:physics-captured}).
Contact coverage is the fraction of reference contact frames where the labeled hand lies within 2\,cm of the moving link's contact region with nonzero contact force, using 2,048 sampled surface points on ARCTIC.

\paragraph{Tracking and joint errors.}
G-MPJPE is the mean global position error over the 22 body joints, and AAE is the mean absolute error of the active joint over the reference-active frames, in degrees for revolute and centimeters for prismatic joints.
For the free-root ARCTIC objects, we additionally report the mean world-frame translation error and the mean geodesic rotation error of the object root.

\paragraph{Reconstruction metrics.}
\label{app:recon-interaction-metrics}
For each annotated contact frame and fingertip, Contact measures the mean one-way distance from the fingertip-region vertices to the nearest vertices in the object contact region.
For D3D-HOI, we annotated the contact frames and the object contact regions.
Pen.\ is the percentage of human vertices inside the articulated object mesh, averaged over frames and videos.
Part IoU is the frame-averaged IoU between the rendered visible mask of the moving part and its input mask.
Body 2D and Hand 2D are the mean pixel distances between the projected SMPL-X body and hand joints and the 2D keypoints detected by ViTPose~\citep{xu2022vitpose} and WiLoR~\citep{potamias2025wilor}.
Human and object smoothness are the mean norm of the per-vertex second-order finite difference across frames.
On generated videos, we follow GRAIL~\citep{xie2026grail} for contact, penetration, and smoothness alongside part IoU and 2D keypoint errors, averaging all metrics per video and then across videos.

\paragraph{Executed-motion jerk.}
We compute executed-motion jerk as the second finite difference of the simulated linear velocities of the 22 body joints at 30\,Hz, taking the maximum norm over joints at each step and averaging over time and over the 128 evaluated trajectories.

\section{Additional Results}
\label{app:additional-results}

\subsection{Additional Comparisons}

\paragraph{Sensitivity to the progress threshold.}
\cref{tab:progress-threshold} reports the success rate on D3D-HOI as the progress threshold increases from 0.5 to 0.8.
\om{} achieves the highest success at every threshold, decreasing from 68.1\% to 53.4\%, while CoDA decreases from 38.6\% to 25.9\% and InterMimic and HDMI approach zero.
\begin{table}[t]
  \caption{\textbf{Success rate under different progress thresholds on D3D-HOI.}
  Each column reports the success rate (\%) of the per-asset policies
  at a different cutoff $\tau$ for phase-duration-weighted mean progress $P$.
  All columns use the remaining success conditions in \cref{app:metric-audit}.
  $^\dagger$ denotes task adaptation.}
  \vspace{-0.4em}
  \label{tab:progress-threshold}

  \centering
  \small
  \setlength{\tabcolsep}{2.9pt}
  \renewcommand{\arraystretch}{1.08}
  \begin{tabular}{l*{7}{c}}
    \toprule
    \textbf{Method} & $\tau{=}0.50$ & $0.55$ & $0.60$ & $0.65$ & $0.70$ & $0.75$ & $0.80$ \\
    \midrule
    InterMimic$^\dagger$~\citep{xu2025intermimic} & 17.5 & 11.6 & 5.6 & 2.4 & 2.4 & 2.0 & 1.2 \\
    HDMI~\citep{weng2025hdmi} & 4.4 & 2.0 & 1.2 & 0.8 & 0.0 & 0.0 & 0.0 \\
    CoDA~\citep{pi2025coda} & 38.6 & 36.3 & 33.5 & 31.5 & 30.7 & 28.3 & 25.9 \\
    \rowcolor{OursBlue}
    \textbf{Ours} & \textbf{68.1} & \textbf{64.5} & \textbf{60.6} & \textbf{58.6} & \textbf{57.4} & \textbf{55.4} & \textbf{53.4} \\
    \bottomrule
  \end{tabular}
\end{table}

CoDA makes substantial joint progress when each policy is trained on a
single reference (\cref{tab:physics-single,tab:physics-captured}).
\om{} still achieves the highest success rate, joint progress, and contact
coverage on both benchmarks.
\begin{table*}[t]
  \caption{\textbf{Per-trajectory physics execution on D3D-HOI~\citep{xu2021d3dhoi}.}
  We train one policy for each of 45 human--object trajectories reconstructed from real videos.
  In simulation, the humanoid must track the reference body motion and move the passive object joint through contact.
  Success rate (SR) uses an 80\% threshold on phase-duration-weighted joint progress.
  $^\dagger$ denotes task adaptation.}
  \vspace{-0.4em}
  \label{tab:physics-single}

  \centering
  \small
  \setlength{\tabcolsep}{6pt}
  \renewcommand{\arraystretch}{1.08}
  \setlength{\aboverulesep}{1pt}
  \setlength{\belowrulesep}{1pt}
  \begin{tabularx}{\textwidth}{lcc*{4}{>{\centering\arraybackslash}X}}
    \toprule
    \multirow{2}{*}{\textbf{Method}}
      & \multirow{2}{*}{\textbf{SR}~(\%)$\,\uparrow$}
      & \multirow{2}{*}{\textbf{G-MPJPE}~(mm)$\,\downarrow$}
      & \multicolumn{2}{c}{\textbf{AAE}$\,\downarrow$}
      & \multirow{2}{*}{\textbf{Prog.} (\%)$\,\uparrow$}
      & \multirow{2}{*}{\textbf{Cov.} (\%)$\,\uparrow$} \\
    \cmidrule(lr){4-5}
      & & & Rev. ($^\circ$) & Pris. (cm) & & \\
    \midrule
    InterMimic$^\dagger$~\citep{xu2025intermimic} & 2.9 & 285.8 & 28.38 & 13.20 & 48.9 & 52.3 \\
    HDMI~\citep{weng2025hdmi} & 5.6 & 595.5 & 28.81 & 11.33 & 58.2 & 4.4 \\
    CoDA~\citep{pi2025coda} & 64.1 & 345.6 & 8.84 & 9.71 & 82.7 & 53.1 \\
    \rowcolor{OursBlue}
    \textbf{Ours} & \textbf{90.5} & \textbf{84.4} & \textbf{5.13} & \textbf{3.21} & \textbf{94.2} & \textbf{85.9} \\
    \bottomrule
  \end{tabularx}
\end{table*}

\begin{table*}[t]
  \caption{\textbf{Per-trajectory physics execution on ARCTIC~\citep{fan2023arctic}.}
  We train one policy for each of 40 motion-captured human--object trajectories.
  The objects can move freely, so we also report object-root translation (Root-T) and rotation (Root-R) errors.
  Success rate (SR) uses an 80\% threshold on phase-duration-weighted joint progress.
  $^\dagger$ denotes task adaptation.}
  \vspace{-0.4em}
  \label{tab:physics-captured}

  \centering
  \small
  \setlength{\tabcolsep}{6pt}
  \renewcommand{\arraystretch}{1.08}
  \setlength{\aboverulesep}{1pt}
  \setlength{\belowrulesep}{1pt}
  \begin{tabularx}{\textwidth}{lcc*{5}{>{\centering\arraybackslash}X}}
    \toprule
    \multirow{2}{*}{\textbf{Method}}
      & \multirow{2}{*}{\textbf{SR}~(\%)$\,\uparrow$}
      & \multirow{2}{*}{\textbf{G-MPJPE}~(mm)$\,\downarrow$}
      & \multicolumn{3}{c}{\textbf{Object errors}$\,\downarrow$}
      & \multirow{2}{*}{\textbf{Prog.} (\%)$\,\uparrow$}
      & \multirow{2}{*}{\textbf{Cov.} (\%)$\,\uparrow$} \\
    \cmidrule(lr){4-6}
      & & & AAE ($^\circ$) & Root-T (cm) & Root-R ($^\circ$) & & \\
    \midrule
    InterMimic$^\dagger$~\citep{xu2025intermimic} & 0.0 & \textbf{104.8} & 53.3 & \textbf{16.7} & 51.1 & 47.0 & 38.1 \\
    HDMI~\citep{weng2025hdmi} & 0.0 & 278.8 & 58.2 & 64.2 & 96.2 & 49.1 & 7.4 \\
    CoDA~\citep{pi2025coda} & 32.5 & 779.3 & 37.4 & 35.2 & \textbf{43.8} & 63.8 & 4.2 \\
    \rowcolor{OursBlue}
    \textbf{Ours} & \textbf{62.5} & 149.5 & \textbf{36.3} & 20.1 & 48.5 & \textbf{82.8} & \textbf{47.2} \\
    \bottomrule
  \end{tabularx}
\end{table*}

\begin{figure*}[p]
  \centering
  \includegraphics[width=\textwidth,height=0.92\textheight,keepaspectratio]{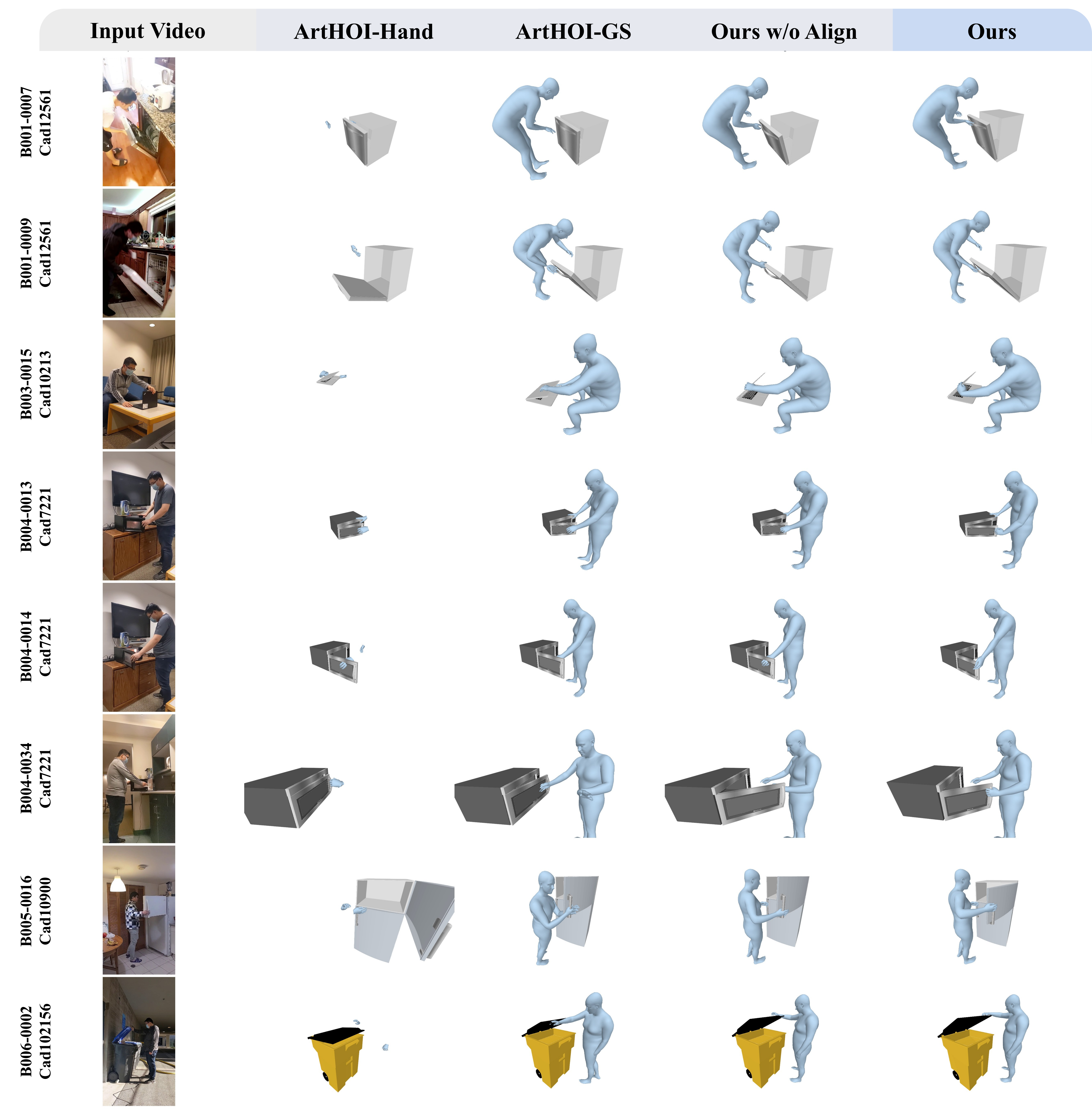}
  \caption{\textbf{Additional reconstruction comparisons (cases 1--8).} Each row shows the input and reconstructions at the selected contact frame.}
  \label{fig:reconstruction-contact-grid-page1}
\end{figure*}

\begin{figure*}[p]
  \centering
  \includegraphics[width=\textwidth,height=0.92\textheight,keepaspectratio]{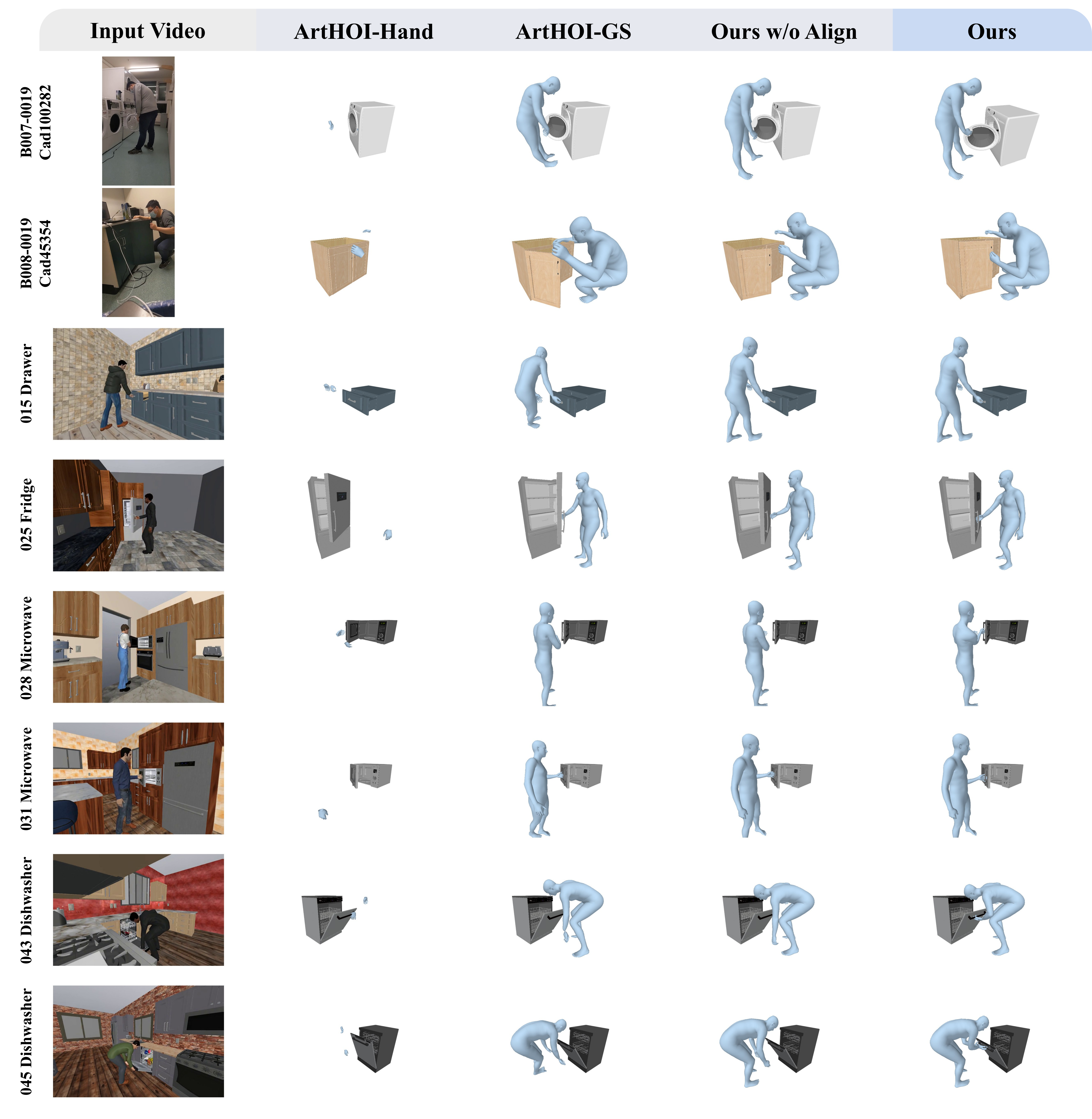}
  \caption{\textbf{Additional reconstruction comparisons (cases 9--16).} Each row shows the input and reconstructions at the selected contact frame.}
  \label{fig:reconstruction-contact-grid-page2}
\end{figure*}

\Cref{fig:reconstruction-contact-grid-page1,fig:reconstruction-contact-grid-page2} provide additional qualitative comparisons on sixteen D3D-HOI and RoboCasa cases. Each row shows the input frame, ArtHOI-Hand, ArtHOI-GS, Ours without alignment, and Ours.

\subsection{Computational Cost}
\label{app:runtime}

\paragraph{Reconstruction.}
\Cref{tab:runtime} reports the mean preprocessing time per video on D3D-HOI.
The timed stages take 9.5 minutes per video on average.
Mask segmentation, human motion estimation, depth prediction, and point tracking account for 96.3\% of this time.
\begin{table}[t]
  \caption{\textbf{Reconstruction preprocessing time on D3D-HOI.}
  Mean wall-clock times on one GPU per video, measured over 253 D3D-HOI videos of 50--302 frames with complete stage timings.}
  \vspace{-0.4em}
  \label{tab:runtime}

  \centering
  \small
  \begin{tabular*}{\linewidth}{l@{\extracolsep{\fill}}c}
    \toprule
    Stage & Time (s) \\
    \midrule
    Mask segmentation (SAM 2 and SAM 3) & 178.9 \\
    Depth prediction (Depth Anything 3) & 139.5 \\
    Human motion estimation (SAM 3D Body) & 140.2 \\
    Human--depth alignment & 5.8 \\
    Point tracking (CoTracker3) & 89.0 \\
    Object pose and scale initialization & 13.6 \\
    Active-joint selection & 1.6 \\
    \midrule
    \textbf{Total preprocessing time} & \textbf{568.6} \\
    \bottomrule
  \end{tabular*}
\end{table}

\paragraph{Physics refinement.}
We train one residual RL policy per CAD asset using its reference trajectories.
Each policy is trained with 4,096 parallel environments for 1,000 updates,
collecting 32 steps per environment per update for a total of 131.1 million
environment steps per policy.

\subsection{Object-Goal Policy Distillation}
\label{app:student-diagnostics}

\paragraph{Student training.}
For each articulated asset, we distill its reference-conditioned physics teacher into a student that observes the current humanoid and object state and future active-joint targets at offsets of 1, 4, 16, and 64 control frames, together with their differences from the current joint state.
The student receives neither future human poses nor the teacher action at inference and outputs a full-body PD target.
We adapt the online distillation framework of InterMimic~\citep{xu2025intermimic}, in which rollouts gradually switch from teacher to student actions while the teacher labels the visited states, and combine action supervision with reinforcement learning on the tracking, contact, and articulation-progress rewards of the teacher.

\paragraph{Evaluation.}
We evaluate each reference once without perturbation and ten times per perturbation setting, with identical initial states for the Human and Human + Goal settings.
Goal perturbations add uniform noise at knots one second apart, bounded by 10$^\circ$, which fades in over the first second and is attenuated near joint limits.

\begin{figure}[tp]
  \centering
  \includegraphics[width=\linewidth]{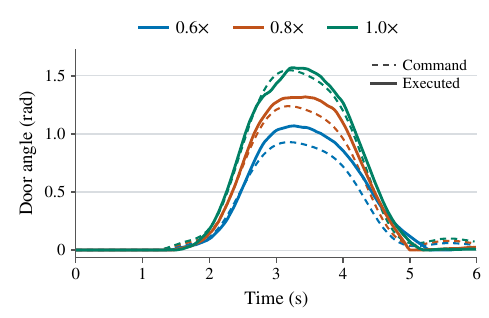}
  \caption{\textbf{Microwave student response to door-angle commands.}
  We vary the command amplitude for one training reference and keep the initial state fixed.
  Dashed curves show the commanded angle, and solid curves show the executed angle.}
  \vspace{-0.4em}
  \label{fig:student-goal-response}
\end{figure}

A student rollout succeeds if it reaches the full horizon, achieves at least 50\% progress in each phase of the perturbed command, and satisfies shared kinematic smoothness thresholds across body velocity, acceleration, jerk, and contact-foot motion.
AAE measures how closely the executed motion follows the perturbed command over the valid prefix, and non-foot low-body contacts, judged from the net body force and the distance to the active object link, count as falls.
In \cref{fig:student-goal-response}, the microwave student opens and closes the door at all three command amplitudes, and the peak opening angle increases with the commanded amplitude.

\subsection{Cross-Embodiment Transfer to a Unitree G1}
\label{app:g1-application}

\begin{figure}[tp]
  \centering
  \includegraphics[width=0.95\linewidth]{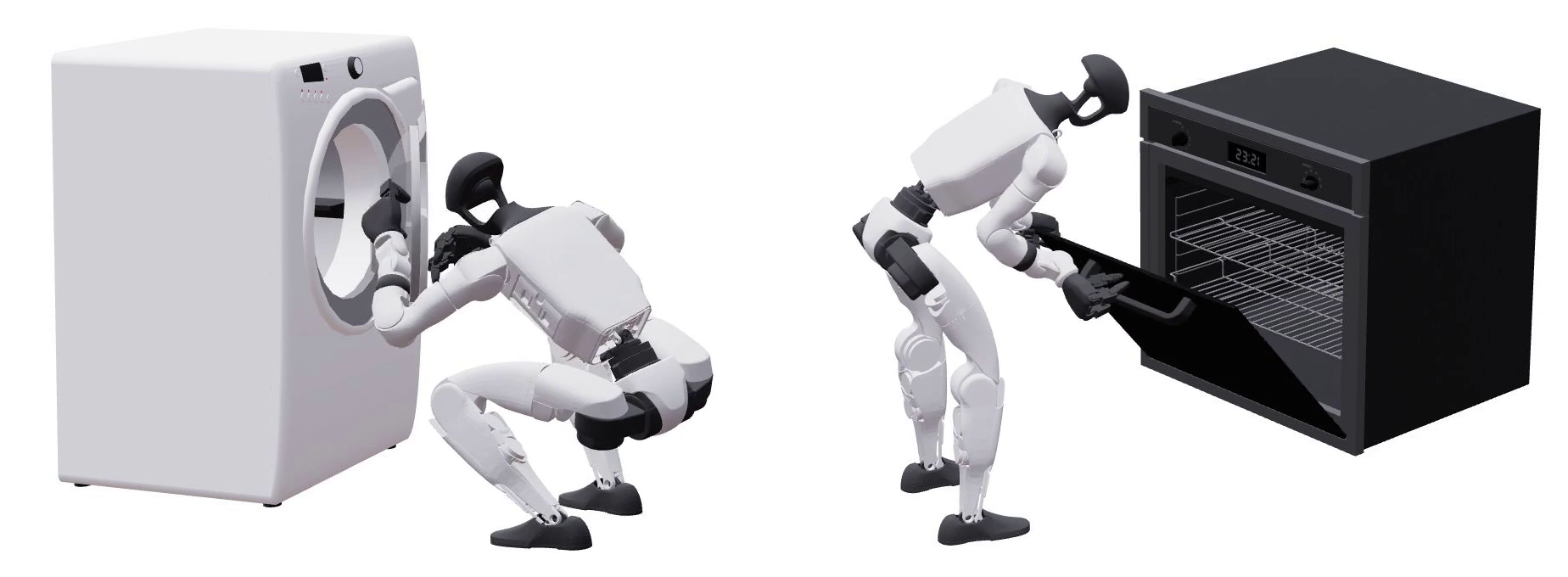}
  \caption{\textbf{Cross-embodiment transfer to a Unitree G1.}
  The robot follows retargeted human--object trajectories in simulation,
  pulling open the side-hinged door of a washing machine (left) and lowering an oven door by its handle (right).}
  \vspace{-0.4em}
  \label{fig:g1_hoi}
\end{figure}

We retarget the reconstructed trajectories to a Unitree G1~\citep{unitreeg1} with the SUGAR~\citep{wu2026sugar} codebase and adapt the rewards of our physics refiner to the G1 embodiment.
\cref{fig:g1_hoi} shows two keyframes from the simulated rollouts (videos are available on our \href{\projecturl}{project page}).

\subsection{Failure Analysis}
\label{app:failure-analysis}

\begin{figure}[!htp]
  \centering
  \includegraphics[width=\linewidth]{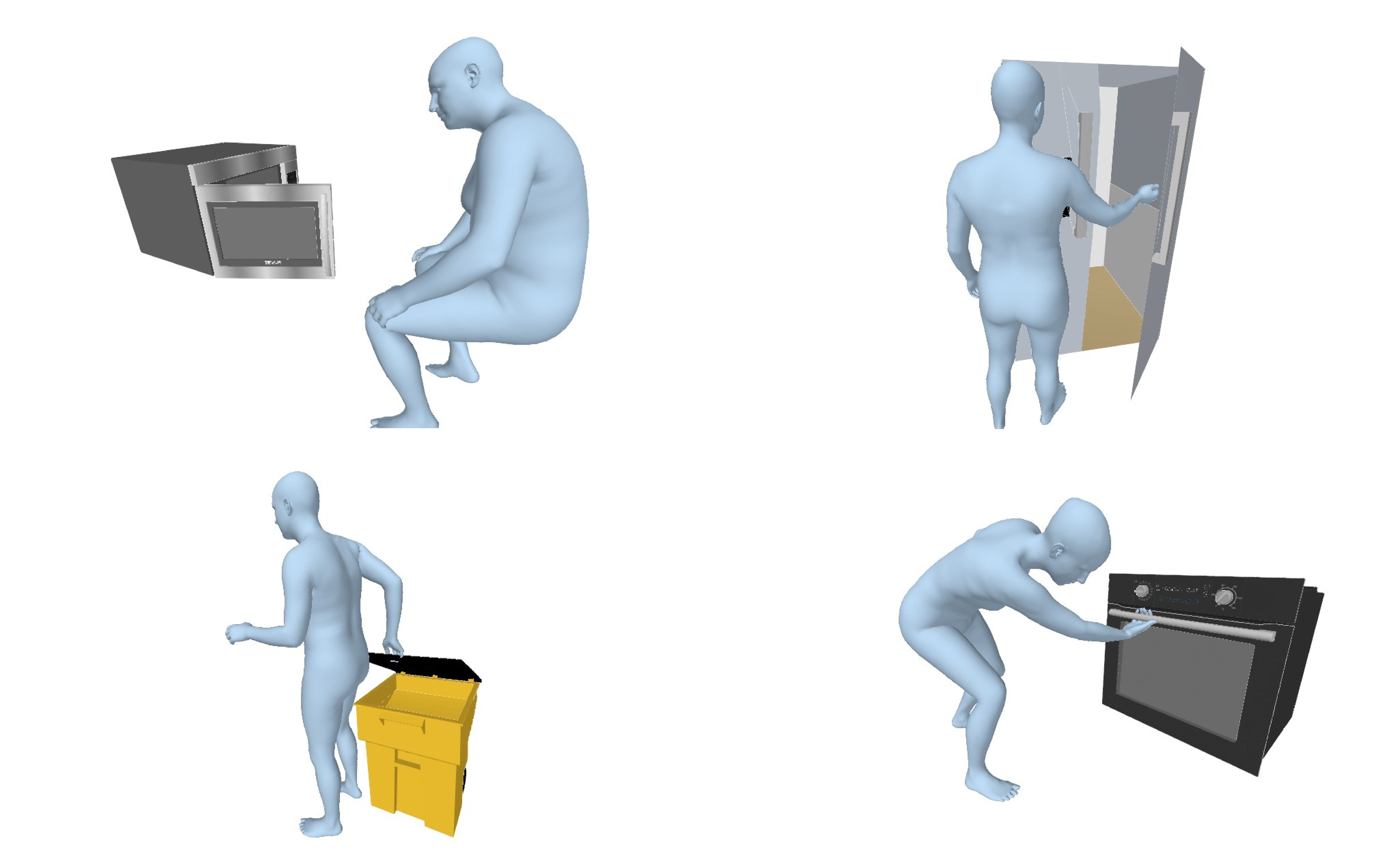}
  \caption{\textbf{Examples of failure modes.}
  The snapshots illustrate failures to establish or maintain effective contact and failures to actuate the object after contact.}
  \vspace{-0.4em}
  \label{fig:failure-analysis}
\end{figure}

We group failed physics-refinement rollouts into two typical categories (\cref{fig:failure-analysis}). \textit{(1) Contact acquisition and maintenance failures.} The hand either fails to reach the intended contact region or loses contact before completing the interaction, resulting in little object motion. \textit{(2) Object interaction and actuation failures.} Despite establishing contact, the hand fails to produce the intended object motion due to spatial misalignment caused by object reconstruction errors, insufficient force along the required direction, or failure to sustain the interaction. In both cases, the articulated object makes limited progress toward its target state.

\begin{figure*}[p]
  \centering
  \begin{tcolorbox}[
    enhanced,
    width=\linewidth,
    colback=promptbg,
    colframe=promptframe,
    arc=2mm,
    boxrule=0.8pt,
    left=7pt, right=7pt, top=9pt, bottom=7pt,
    fontupper=\rmfamily\fontsize{9.3}{11.3}\selectfont,
    fonttitle=\rmfamily\large\bfseries,
    title={Prompts for Finger-Contact Labeling},
    coltitle=white,
    attach boxed title to top center={yshift=-2mm},
    boxed title style={colback=promptframe, rounded corners}
  ]
    \tcbset{prompt output/.style={
      colback=white, colframe=promptframe!23, sharp corners, boxrule=0.5pt,
      boxsep=0pt, left=6pt, right=6pt, top=4pt, bottom=4pt,
      before skip=5pt, after skip=0pt
    }}
    \newcommand{\PromptMono}{\fontencoding{T1}\fontfamily{lmtt}\selectfont}
    \newcommand{\PromptDivider}{\par\vspace{5pt}{\color{promptframe!35}\hrule height 0.35pt}\vspace{5pt}}
    \SectionTitle{Stage 1: Whole-Video Prompt}

    The following are timestamped, uniformly sampled full-scene images spanning the entire video.
    Identify the articulated object and its moving part, and describe how it changes over time.
    For each time interval, assess whether the person's anatomical left and/or right hand actively moves the articulated part or maintains its manipulated position.
    A hand resting on the fixed body, holding a separate object, or merely nearby is not active.
    A hand maintaining a grip during a brief pause can be active.
    Both hands, hand switches, and no active hand are possible.
    Do not infer handedness from screen side alone; mark uncertain when the anatomy, contact, or role is unclear.
    Sparse images can miss short events; report uncertainty and do not invent precise boundaries.

    \begin{tcolorbox}[prompt output]
      \SubTitle{Output format}\\[0.4ex]
      \fontsize{8.4}{10.1}\PromptMono
      \{"target\_object": "...", "motion": "...",\\
      \hspace*{1em}"intervals": [\{"start\_seconds": 0.0, "end\_seconds": 1.0,\\
      \hspace*{2em}"left": "active/passive/none/uncertain", "right": "active/passive/none/uncertain",\\
      \hspace*{2em}"evidence": "short description"\}]\}
    \end{tcolorbox}

    \PromptDivider
    \SectionTitle{Stage 2: Per-Frame Prompt}

    You are given PREVIOUS, TARGET, and NEXT frames from a video.
    Each time has a full-scene image and labelled LEFT/RIGHT HMR-projected hand crops when available.
    Return contact labels for the TARGET frame only.

    \par\vspace{5pt}
    \SubTitle{Target object}\enspace
    First identify the target articulated object: the object whose connected part moves relative to its main body as the person operates it.
    Use the full-scene images to distinguish this object and its moving part from any separate objects nearby or held by the person.
    Do not count contact with a separate item, the environment, the person's body, or the target's fixed body as contact with its movable part.

    \par\vspace{5pt}
    \SubTitle{Active hand}\enspace
    For each labelled hand, determine whether it is an active manipulation hand.
    A hand is \textbf{active} only if it touches the target's moving part or its attached handle/control and contributes to moving that part or maintaining its manipulated position.
    A hand merely holding another item, resting on the fixed body, or near the object is not active.
    A hand that maintains a grip can remain active during a brief pause; it does not need to move at every frame.
    Do not infer active manipulation from action category, proximity, or image-box overlap alone.

    \par\vspace{5pt}
    \SubTitle{Fingers and temporal evidence}\enspace
    For each active hand, decide separately which of the thumb, index, middle, ring and pinky contact the target moving part or attached handle/control.
    Do not assume all five touch.
    For a hand that is passive, not in contact, or uncertain, set every finger false.
    Use the provided LEFT/RIGHT crop labels, which refer to the person's actual hands.
    Use temporal visual evidence for brief occlusion: a visible continuous grip before and after an occlusion can support contact through it; a hidden finger alone does not prove release.
    Visible separation or release is evidence against contact.
    If the images do not support an active grip or contact for a particular finger, do not guess: use uncertain for the role or false for that finger.

    \par\vspace{5pt}
    \SubTitle{Hand crops}\enspace
    Hand crops are positioned using HMR body-model projections, not a visual hand detector.
    A crop may be provided even when the hand or fingers are occluded or outside the image.
    The box indicates only an estimated hand location and anatomical side: it does not prove that the hand is visible or touching the object.
    Do not infer contact from box overlap.
    When the hand or fingers are hidden, use prior and subsequent frames and the full action to judge a continuous grip; if evidence is insufficient use uncertain or false rather than guessing.

    \begin{tcolorbox}[prompt output]
      \SubTitle{Output format}\\[0.4ex]
      \fontsize{8.4}{10.1}\PromptMono
      \{"target\_object": "short visual description",\\
      \hspace*{1em}"left": \{"role": "active/passive/none/uncertain",\\
      \hspace*{2em}"fingers": \{"thumb": false, "index": false, "middle": false, "ring": false, "pinky": false\}\},\\
      \hspace*{1em}"right": \{\emph{same fields as left}\}\}
    \end{tcolorbox}

    \PromptDivider
    \SectionTitle{Appended to the Per-Frame Prompt}

    Whole-video sparse-frame assessment (fallible, not ground truth): {\PromptMono\{stage-1 answer\}}.
    Use the interval times only as coarse context.
    The target frame must be judged using its local images; override the global assessment if local evidence disagrees.
    Do not infer contact from the global assessment alone.
    Target times can be obtained by dividing the frame index by {\PromptMono\{fps\}} frames per second.
  \end{tcolorbox}
  \caption{\textbf{Prompts for finger-contact labeling.}
  The first prompt is sent once per video with the sampled full-scene frames, and the second prompt is sent for every queried frame with the images of \cref{fig:contact-visual-prompt} and the answer to the first prompt.}
  \vspace{-1.2em}
  \label{fig:contact-prompt}
\end{figure*}

\section{Limitations and Future Work}
\label{app:limitations}

\paragraph{Occlusion, monocular ambiguity, and thin contact gaps.}
\om{} still fails under severe occlusion and when the contact gap is very thin.
Occluding the hand and occluding the moving part hurt different stages.
When the manipulating hand is heavily occluded, VLM contact labels and finger-surface alignment become unreliable or incomplete.
When the moving part is occluded, point tracks and silhouettes degrade, which impairs articulation tracking.
Depth-scale uncertainty and foreshortening can also make the reconstruction underestimate the joint amplitude, so the physics refiner receives an understated target.
Stronger contact perception and an explicit model of reference uncertainty would let the refiner discount unreliable segments.

\paragraph{Embodiment gap and qualitative G1 scope.}
Our current Unitree G1 results are qualitative demonstrations in simulation.
Because of their size, the G1 hands may also struggle to grasp thin or recessed handles.

\paragraph{Asset requirement and future directions.}
Each input video must be paired with an articulated asset that provides its geometry and kinematics.
Future work includes scaling to more diverse interaction data and articulated assets, for example through asset retrieval and multi-joint or multi-object scenes, running broader quantitative G1 experiments on contact, actuation, and stability in simulation, and then moving to hardware evaluation.

\paragraph{Broader Impact.}
\label{app:broader-impact}
\om{} generates physically executable full-body interactions with articulated objects from video, which can support digital human animation, embodied AI research, and robot manipulation demonstrations.
The same capability could be misused to fabricate or misrepresent human behavior in video, and reconstructing from human videos raises privacy concerns for the people who appear in them.
Our experiments use only publicly available benchmark data and generated videos, and any use on footage of real people should follow the consent and licensing terms attached to it.
The learned policies also inherit the biases of the simulation environment and the assets they are trained on.

\end{document}